\documentclass[a4paper,12pt]{article}
\usepackage[utf8]{inputenc} 
\usepackage[T1]{fontenc}    
\usepackage{PRIMEarxiv}

\let\cite\parencite

\title{Learning collision risk proactively from naturalistic driving data at scale
\thanks{Officially published in \textit{Nature Machine Intelligence}, accessible at
\href{https://doi.org/10.1038/s42256-026-01189-w}{https://doi.org/10.1038/s42256-026-01189-w}}
}

\author{
  Yiru Jiao$^{1,3}$, Simeon C. Calvert$^{1,3}$, Sander van Cranenburgh$^{2,3}$, Hans van Lint$^{1}$ \\
  $^1$ Department of Transport \& Planning,\\ $^2$  Department of Engineering Systems and Services,\\ $^3$ CityAI lab,\\
  Delft University of Technology, Delft, the Netherlands\\ 
}

\begin{document}
\maketitle

\begin{abstract}
Accurately and proactively alerting drivers or automated systems to emerging collisions is crucial for road safety, particularly in highly interactive and complex urban environments. Existing methods either require labour-intensive annotation of sparse risk, struggle to consider varying contextual factors, or are tailored to limited scenarios. Here we present the Generalised Surrogate Safety Measure (GSSM), a data-driven approach that learns collision risk from naturalistic driving without the need for crash or risk labels. Trained over multiple datasets and evaluated on 2,591 real-world crashes and near-crashes, a basic GSSM using only instantaneous motion kinematics achieves an area under the precision-recall curve of 0.9, and secures a median time advance of 2.6 seconds to prevent potential collisions. Incorporating additional interaction patterns and contextual factors provides further performance gains. Across interaction scenarios such as rear-end, merging, and turning, GSSM consistently outperforms existing baselines in accuracy and timeliness. These results establish GSSM as a scalable, context-aware, and generalisable foundation to identify risky interactions before they become unavoidable, supporting proactive safety in autonomous driving systems and traffic incident management. Code and experiment data are openly accessible at \url{https://github.com/Yiru-Jiao/GSSM}.

\end{abstract}

\keywords{Road safety \and collision risk \and risk quantification \and trajectory reconstruction \and naturalistic driving data}

\newpage
\section{Introduction}\label{sec: introduction}
\begin{figure}[htbp]
    \centering
    \includegraphics[width=\linewidth]{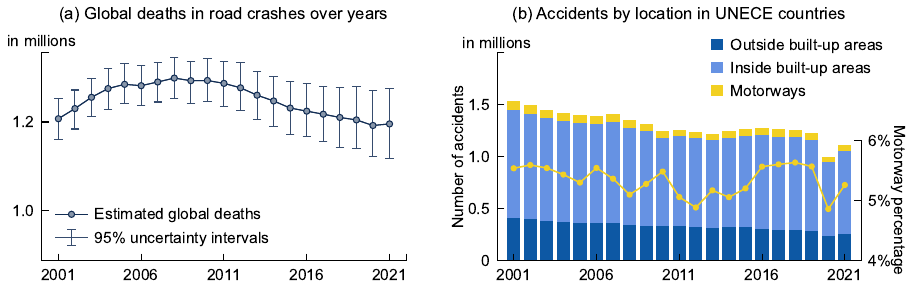}
    \caption{Statistics of traffic accidents. \textbf{(a)}~Estimated numbers of fatalities due to road traffic accidents per year by the Institute for Health Metrics and Evaluation~\cite{GBD2022}. \textbf{(b)}~Distribution of traffic accidents that occurred in different locations from 2001 to 2021 in 27 countries where complete data are accessible. The data is sourced from the United Nations Economic Commission for Europe (UNECE) Statistical Database~\cite{dataUNECE}.}
    \label{fig: Global_crashes}
\end{figure}

Road traffic safety remains a critical global concern. Over one million fatalities were recorded on roads due to traffic accidents worldwide every year, as well as ten times as many injuries~\cite{WHO2023}. With advancements in vehicle safety technologies and policy improvements, a significant number of fatalities have been reduced since 2010. However, this reduction has plateaued in recent years, as evidenced by Figure~\ref{fig: Global_crashes}(a). Notably, fewer than 6\% of all traffic accidents occur on high-speed motorways, as displayed in Figure~\ref{fig: Global_crashes}(b). Instead, the majority of crashes occur on urban roads, where the traffic situation is more complex due to various types of road users and their multi-directional interactions. Given that 60\% of the global population is expected to reside in urban areas by 2030~\cite{UNESA2020}, improving traffic safety in highly interactive urban environments has become both urgent and globally relevant.

A core challenge lies in accurately quantifying collision risk in real time to allow for prevention before crashes happen. Traffic safety research is traditionally grouped into two categories, reactive and proactive, based on whether the focus is on ex post analysis or ex ante prevention. Reactive research identifies risk factors from real-world crashes to derive improvements in road topology design and infrastructure safety monitoring~\cite{Saul2021} (see~\cite{Papadimitriou2019,Jakobsen2022} for relevant literature reviews). Relying on historical crash data, such improvements are inevitably delayed after injuries and damage have occurred. In contrast, proactive research anticipates potential collisions to enable timely interventions~\cite{Horrey2012,Chang2017}. This means being alert to the precursors to collisions, such as near misses, hazardous interactions, and evolving traffic conflicts, which potentially lead to crashes if not mitigated.

To address this challenge of proactive collision risk quantification, several methodologies have been explored in different fields. The earliest established methodology is surrogate safety measures (SSM), also referred to as surrogate measures of safety (SMoS) or criticality metrics within autonomous driving research~\cite{Arun2021ssmreview,Wang2021,Westhofen2022metricreview}. Developed over decades since the 1970s~\cite{conflict1977,Hayward1972ttc,Cooper1976drac}, SSMs estimate the likelihood or severity of potential collisions through physics-based indicators designed for specific road user behaviour such as car-following and lane-changing. The underlying assumption is that crash risk arises when the current situation will result in a collision unless immediate evasive actions are taken. Well-known examples include Time-to-Collision (TTC), deceleration rate to avoid collision (DRAC), artificial potential fields (e.g., safety field~\cite{Wang2015field} and risk field~\cite{MullakkalBabu2020,Kolekar2020}), and other varying spatiotemporal measures. Although these indicators are intuitive, traffic collisions are not the result of a certain type of road user behaviour, but involve various types of interactions in diverse and dynamic traffic environments~\cite{EC2024}. Therefore, the specifically designed SSMs lack adequate \textbf{\emph{context-awareness}}~\cite{Dey2001} to consider heterogeneous interactions, different types of road users, and broader environment factors. This lack also limits the \textbf{\emph{generalisability}} of SSMs beyond their designed conditions. 

A methodology widely used in robotics and the control domain quantifies collision risk based on motion prediction under uncertainty~\cite{Lefvre2014,Dahl2019}. It predicts the future positions or reachable sets of road users, propagates uncertainty through these predictions, and assesses probabilistic violations of safety constraints (e.g.,~\cite{Althoff2014,Kim2018,mathiesen2024}). Risk quantification in this methodology implicitly assumes that the uncertainties of road user behaviour can be sufficiently captured by a stochastic model, which remains valid in safety-critical situations. In practice, this assumption is often violated by aggressive, inattentive, or other anomalous behaviour that typically precedes crashes. To accommodate such more complex uncertainties, recent studies have increasingly integrated data-driven prediction models~\cite{McAllister2017,Li2024}. While these models enhance context-awareness as traffic environments are incorporated, the challenge of model \textbf{\emph{generalisability}} persists in intensely interactive and safety-critical scenarios that are under-represented in normal training data. When encountering such scenarios, the accuracy and reliability of risk quantification may degrade significantly. 

Facilitated by the rapid development of computer vision, a video-data-driven methodology has emerged since 2018 known as Traffic Accident Anticipation (TAA, see dedicated literature reviews such as~\cite{Kataoka2018,Fang2024taareview}). Its objective is to provide early predictions of impending collisions by modelling the visual patterns across sequential video frames. This is commonly achieved through supervised learning with real-world accidents serving as labels. The fundamental assumption is the existence of representative visual cues before accidents happen, so that deep neural networks can identify these cues. For example, vehicles deviating from their lanes may signal loss of control or evasive manoeuvres~\cite{Patera2025}, while motorcycles at high speed can indicate risk in urban areas~\cite{Kumamoto2025}. Heavily driven by annotated crash data, TAA faces a practical challenge of \textbf{\emph{scalability}}. The infrequency and variety of traffic accidents, of which the videos are inherently difficult to acquire, impede effective and reliable training~\cite{Liu2024}. Similar to the previous methodologies, TAA models also struggle with \textbf{\emph{generalisability}} to new contexts that differ, even slightly, from training patterns.

In summary, to proactively quantify collision risk, existing methodologies share 3 limitations in
\begin{itemize}[noitemsep,topsep=-\parskip] 
    \item \textbf{scalability}, leveraging large-scale everyday observations instead of relying on factual crashes;
    \item \textbf{context-awareness}, accounting for any information that can be used to characterise interaction~\cite{Dey2001} in addition to road user behaviour, such as weather, lighting, and road conditions; and
    \item \textbf{generalisability}, covering a wide variety of traffic interactions from on motorways to at urban intersections, and handling new contexts not seen in training data.
\end{itemize}

In this paper, we propose the generalised surrogate safety measure (GSSM), a novel approach to proactive risk quantification of potential traffic collisions. GSSM evaluates how extreme a given traffic interaction deviates from typical safe behaviour towards an unsafe state in the interaction context, outputting a continuous risk level and the corresponding likelihood of a potential collision. This consideration of extreme interaction makes it possible to naturally generalise from normal interactions to safety-critical situations. GSSM utilises neural networks to enable context-awareness, incorporating various relevant contextual information (e.g., motion states of road users, weather, and road conditions) and thus adapting to diverse scenarios from vehicle-vehicle encounters on highways to pedestrian-vehicle interactions in urban streets. Furthermore, GSSM learns interaction patterns from naturalistic data, i.e., real-world and unconstrained driving behaviour collected under everyday conditions with minimal interference, and does not require any crash records or manually labelled risk. This makes the training of GSSM scalable. In summary, GSSM offers a new paradigm for proactive collision risk quantification. It promises a contribution to advancing proactive safety research, to improve, e.g., autonomous driving systems, infrastructure design and operations, as well as traffic management policies.

In the rest of this paper, Section~\ref{sec: gssm} first defines the problem of proactive collision risk quantification and then introduces GSSM. Its learning is explained in Section~\ref{sec: gssm learning}. In Section~\ref{sec: data}, we present a real-world dataset of crashes and near-crashes for validating GSSM. Our experiment design is described in Section~\ref{sec: exps}, followed by Section~\ref{sec: results} where the characteristics of GSSM are demonstrated. Finally, Section~\ref{sec: conclusion} concludes this paper and envisions future research. To improve readability, we place details that do not hinder understanding in appendices.

\section{Generalised surrogate safety measure} \label{sec: gssm}
In this section, we introduce the generalised surrogate safety measure (GSSM). We first formulate the general problem of proactive collision risk quantification, and then outline the theoretical basis for GSSM. Our design of GSSM inherits the fundamental knowledge developed over decades of research on surrogate safety measures (SSMs), where potential collisions are also called traffic conflicts. This distinguishes GSSM from existing self-supervised (and unsupervised) approaches, which are tailored for computer vision and primarily based on anomaly detection. Despite that, GSSM is adequately flexible to incorporate vision context.

\subsection{Problem formulation} \label{sec: problem}
Although the consequences of traffic accidents vary, this paper considers that every collision should be avoided and thus focuses on the likelihood of potential collisions. Within this scope, we define the proactive risk quantification of potential collisions as a function $Q:\mathcal{S}\mapsto L$, where $\mathcal{S}$ represents a scenario of traffic interaction and $Q(\mathcal{S})$ estimates the likelihood $L$ of an impending collision in this scenario. $L$ is typically expressed as a probability $p\in(0,1)$, or as a numerical or categorical level $M$ that can be monotonically mapped to $p$. 

The primary objective of $Q$ is to accurately and timely foresee a potential collision. Inaccurate and ill-timed anticipation, e.g., road users acting too quickly or, on the contrary, failing to act, are the most common factors of road crashes (as surveyed in~\cite{thomas2013}, 51\% of car drivers, 42\% of motorcyclists, 68\% of pedestrians, and 46\% of cyclists). To alert to a potential collision, $L$ is usually converted into a binary outcome using a threshold. When annotated data of (near-)crashes are available, assessing the accuracy of $Q$ then compares these binary classification results with ground truth annotations, while timeliness is assessed by the time between the threshold-based alerts and the occurrence of a crash or near-crash event. In the absence of annotations, surrogates such as close distance and hard braking are used to create near-crash labels (e.g.,~\cite{Wang2015,Nahata2021,Chen2024}).

As reviewed in Section~\ref{sec: introduction}, $Q$ can be predefined in closed-form expressions or based on uncertainty-aware prediction, and can also be learnt in a supervised manner with annotated crash videos. A particular challenge in collision risk quantification is that crash and near-crash data are expensive to collect and inherently rare, despite causing numerous fatalities and injuries. Consequently, there is a growing interest in unsupervised and self-supervised learning approaches that do not rely on annotated crash data, to name a few, see~\cite{yao2019A3D,Fang2022,yao2022DoTA,Li2023}.

\subsection{Operational definition}\label{sec: model}
\begin{figure}[htbp]
    \centering
    \includegraphics[width=\linewidth]{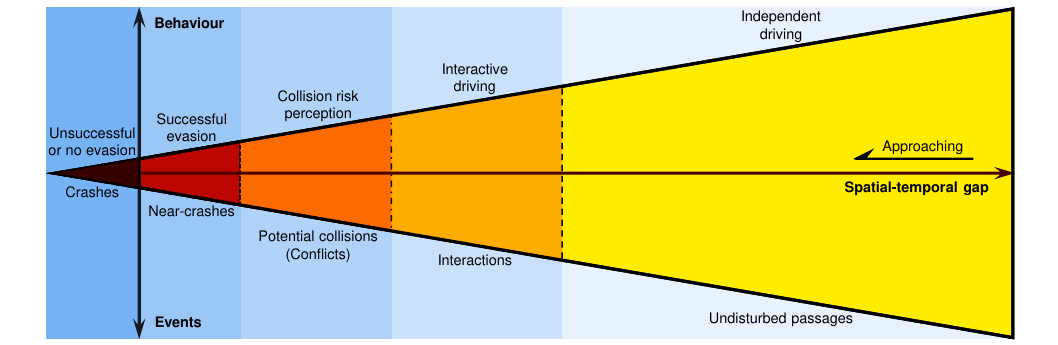}
    \caption{The safety pyramid conceptualises the evolution from safe interactions to unsafe interactions up to crashes.}
    \label{fig: safety_pyramid}
\end{figure}

Every collision evolves, often rapidly, from a previously safe interaction. This evolution was first articulated by Hyd{\'e}n~\cite{Hyden1987ttc} as the safety pyramid. In Figure~\ref{fig: safety_pyramid} we depict the conceptual relations in the safety pyramid, which also reflects the interaction behaviours of road users~\cite{laureshyn2010tadv}. For two or more approaching road users, a potential collision emerges when they are too close to safely interact. The perceived risk of a potential collision then triggers the road users to take evasive actions. In case of successful evasion, the potential collision will end in a near-crash; otherwise, a crash will occur.

Building on the safety pyramid, GSSM considers the spatial-temporal gap between road users as a proxy that can reflect collision risk across varying interaction contexts. Intuitively, a smaller gap implies less physical space and time available to react and prevent a potential collision. Empirical evidence also shows that people perceive increased risk when other moving objects approach~\cite{Schiff1979,Teigen2005}; and motivated by the perceived risk, maintain acceptable separation from each other~\cite{Camara2020}. In this paper, we let GSSM use the spatial gap, hereafter referred to as spacing. GSSM quantifies collision risk by the extent to which a spacing is too close for safe interaction in its interaction context. To enable cross-context risk quantification, we establish two key assumptions:
\begin{description}
    \item[Assumption 1] Potential collisions are precursors to collisions as well as continuation of previously safe interactions.
    \item[Assumption 2] In the same interaction context, reduced spacing monotonically indicates increased risk of collision.
\end{description}

Formally, we consider repeated observations in the same interaction context and define the extent of proximity as the number of times a given spacing is the minimum among all observed spacings. Following the problem formulation, we factorise an interaction scenario $\mathcal{S}$ into $(s,X)$, where $s\geq 0$ is the spacing between two or more road users and $X$ denotes all other contextual observables, e.g., motion states, weather, lighting, and road surface conditions. Accordingly, $\mathrm{GSSM}:(s, X)\mapsto M$, where $M$ is a risk level. For $n\in\mathbb{N}_{>0}$ observations with varying spacings in context $X$, the extreme value theory calculates $p(s,n|X)=\left(\Pr(S>s|X)\right)^n$ as the probability that $s$ is the minimal spacing in $n$ repeated observations. The smaller $s$ is, over the more observations $s$ remains minimal, and the larger $n$ is. 

We interpret the extent of spacing proximity $n$ as conflict intensity. Accordingly, $p(s,n|X)$ can be understood as the probability that an interaction scenario $(s,X)$ is a conflict at intensity $n$. Then we define the critical intensity of an interaction as the smallest $n$ for which the probability of conflict exceeds the probability of no conflict, i.e., $n$ satisfies $p(s,n|X)>1-p(s,n|X)$. Solving the inequality yields $n>\ln0.5/\ln(\Pr(S>s|X))$, and we use this critical point to define GSSM.

To summarise, GSSM is operationally defined as the critical conflict intensity $\hat{n}$ of an interaction scenario $(s,X)$. From the view of extreme value theory, the spacing $s$ has a probability of 0.5 to be the minimum in $\hat{n}$ observations in the same interaction context $X$; from a more intuitive view, if the interaction involves a conflict, its intensity is at least $\hat{n}$. The value of $\hat{n}$ may be very large, e.g., $1,000$ or $10,000$, in high-risk interactions. For convenience, we take the base-10 logarithm of $\hat{n}$, yielding representative scores of 3 and 4, respectively. Equation~\eqref{eq: gssm} presents the complete definition of GSSM with the following properties. Let $M$ be the risk level given by GSSM, $M\in\mathbb{R}$. When $s$ is larger than the median spacing in context $X$, $\Pr(S>s|X)<0.5$ and $M\leq0$, which implies safety. In contrast, smaller $s$ corresponds to larger $M$ and a higher risk of potential collision. Particularly, when $s=0$, $M$ is infinite and indicates a factual collision. In addition, $M$ can be naturally mapped to a probability $p\in(0,1)$ by $p(s,X,M)=\Pr(S>s|X)^{10^M}$. 
\begin{equation}\label{eq: gssm}
    \mathrm{GSSM}(s,X)=\log_{10}\left[\frac{\ln0.5}{\ln(\Pr(S>s|X))}\right]
\end{equation}

In theory, GSSM can be reduced to any SSM that is based on the spatial-temporal gap between road users, by specifying appropriate contextual observables. For example, Time-to-Collision (TTC) and deceleration rate to avoid collision (DRAC) account for the relative speed between two vehicles following one another; Proportion of Stopping Distance (PSD) considers the squared speed of a vehicle following another. For more details on such reduction and additional theoretical derivation, please refer to our previous work~\cite{jiao2024unified}.

\subsection{Multi-directional spacing}\label{sec: 2dspacing}
Spacing serves as a risk proxy in GSSM and reflects the physical and psychological reaction margin to prevent a potential collision. As mentioned in Section~\ref{sec: introduction} and displayed in Figure~\ref{fig: Global_crashes}(b), traffic collisions in urban traffic significantly outnumber those on highways. Urban traffic involves multi-directional movements beyond longitudinal dynamics. These not only include car-following and lane-changing, but also two-dimensional (2D) interactions such as path crossing and conflict negotiation, especially at intersections.

We adapt the method developed in our previous work~\cite{jiao2023inferring} to quantify multi-directional spacing in 2D interactions. Consider two road users, $i$ and $j$. A relative coordinate system is introduced with its origin at the position of $i$ and the y-axis oriented along the direction of the relative velocity $\boldsymbol{v}_{ij}=\boldsymbol{v}_i-\boldsymbol{v}_j$. If $i$ and $j$ have identical velocities, the y-axis is instead aligned with the heading direction of $i$, denoted by $\boldsymbol{h}_i$. We transform the position of $j$, given by $(x_j, y_j)$, into the relative coordinate system and get $(x_{ij}, y_{ij})$. Then the multi-directional spacing between $i$ and $j$ at each time step is represented by $(x_{ij}, y_{ij}, |\boldsymbol{v}_{ij}|)$. This coordinate transformation is explicitly defined in equation~\eqref{eq: 2d spacing}.
\begin{equation}\label{eq: 2d spacing}
\begin{aligned}
    \left[\begin{matrix}x_{ij}\\y_{ij}\end{matrix}\right]&=\frac{1}{\sqrt{x_{\mathrm{axis}}^2+y_{\mathrm{axis}}^2}}\left[\begin{matrix}y_{\mathrm{axis}}&-x_{\mathrm{axis}}\\x_{\mathrm{axis}}&y_{\mathrm{axis}}\end{matrix}\right]+\left(\left[\begin{matrix}x_j\\y_j\end{matrix}\right]-\left[\begin{matrix}x_i\\y_i\end{matrix}\right]\right);\\
    (x_{\mathrm{axis}}, y_{\mathrm{axis}})&=
    \begin{cases}
    (x_{\boldsymbol{v}_i}-x_{\boldsymbol{v}_j},y_{\boldsymbol{v}_i}-y_{\boldsymbol{v}_j}),&\text{if }\boldsymbol{v}_i\neq \boldsymbol{v}_j,\\
    (x_{\boldsymbol{h}_i}, y_{\boldsymbol{h}_i}),&\text{otherwise.}
    \end{cases}
\end{aligned}
\end{equation}

To facilitate subsequent model training, we convert the positional components $(x_{ij}, y_{ij})$ in multi-directional spacing into polar coordinates. We define the polar coordinate system with its pole at the origin of the relative coordinate system and its polar axis aligned with the x-axis of the relative coordinate system. This conversion yields $(\rho_{ij}, s_{ij}, |\boldsymbol{v}_{ij}|)$ as the polar representation of multi-directional spacing, where $\rho_{ij}$ is the angular coordinate and $s_{ij}$ is the radial coordinate. Respectively, $\rho_{ij}$ represents the direction of spacing and $s_{ij}$ represents the distance of spacing. We then incorporate $\rho_{ij}$ and relative speed $|\boldsymbol{v}_{ij}|$ in the context $X$, and use $s_{ij}$ as the spacing variable $s$.

\subsection{Parameterised GSSM}\label{sec: prior distribution}
We now introduce a parametric form of GSSM. Notice that $\Pr(S>s|X)$ in equation~\eqref{eq: gssm} equals to $1-\Pr(S\leq s|X)$. Consider the conditional distribution of spacing $s$ in context $X$, $\Pr(S\leq s|X)$ is the cumulative probability distribution of $p(s|X)$. We thus parameterise $\Pr(S\leq s|X)$ by $F_S(s;\phi(X))$, where $\phi(X)$ denotes the conditional parameters depending on $X$. To specify $F_S(s;\phi(X))$, we can learn $\phi(X)$ using standard statistical and machine-learning techniques, thereby making GSSM more tractable for training.

Many studies demonstrate that the spatial gaps between road users are effectively characterised by the lognormal distribution~\cite{Meng2012,Pawar2016,Anwari2023}. We thus assume the context-conditioned spacing distribution $p(s|X)$ follows a lognormal distribution with two parameters $\mu$ and $\sigma$. For numerical stability in training, we learn $(\mu,\log(\sigma^2))$ instead of $(\mu,\sigma)$. But considering the convenience of notation, we write the estimated parameters $\hat{\phi}(X)=(\hat{\mu}(X), \hat{\sigma}^2(X))$ in this paper. Then equation~\eqref{eq: lognormal pdf} presents the corresponding probability density function, and a parameterised GSSM is defined in equation~\eqref{eq: parameterised gssm}, where $\mathrm{erf}(z)=2\int_0^ze^{-x^2}\mathrm{d}x/\sqrt{\pi}$ is the Gaussian error function within domain $(-1,1)$.
\begin{equation}\label{eq: lognormal pdf}
    f_S(s;\phi(X))=\frac{1}{s\sqrt{2\pi\hat{\sigma}^2(X)}}\exp\left[-\frac{(\ln s-\hat{\mu}(X))^2}{2\hat{\sigma}^2(X)}\right]
\end{equation}

\begin{equation}\label{eq: parameterised gssm}
    \mathrm{GSSM}(s,X)=\log_{10}\left[\frac{\ln0.5}{\ln(1-F_S(s;\phi(X))}\right]=\log_{10}\left[\frac{\ln 0.5}{\ln\frac{1}{2}\left(1-\mathrm{erf}\left(\frac{\ln s-\hat{\mu}(X)}{\sqrt{2\hat{\sigma}^2(X)}}\right)\right)}\right]
\end{equation}

\section{GSSM learning}\label{sec: gssm learning}
For self-supervised risk quantification using GSSM, the primary learning task aims to estimate the conditional parameters of context-conditioned spacing distributions. Notably, learning a GSSM does not require crash or near-crash data; it leverages normal interactions, of which the available data is far more abundant. In addition to this primary task, we introduce two auxiliary tasks. One is context representation learning to incorporate a wider range of data for training; the other is feature attribution to identify contributing factors to potential collision risk.

\subsection{Inference of conditional parameters}\label{sec: inference}
We use a neural network $g_W$ to estimate the conditional parameters $\phi(X)$, where $W$ denotes the network's learnable weights. Therefore, $g_W(X)$ outputs parameter estimates $\hat{\phi}(X)$ and it needs to make $f_S(s;\hat{\phi}(X))$ approximate the spacing distribution $p(s|X)$ as closely as possible. A closer approximation can be indicated by a smaller Kullback–Leibler divergence $D_\mathrm{KL}\left[p(s|X)||f_S(s;g_W(X))\right]$. Minimising this divergence is equivalent to maximising expected log-likelihood $\mathbb{E}_{s\sim p(s|X)}\left[\ln f_S(s;g_W(X))\right]$ given real-world observations of road user interaction. More conveniently, we minimise the negative log-likelihood (NLL) loss defined in equation~\eqref{eq: log likelihood}, where $(s_i,X_i)$ represents a sample of interaction scenario described by spacing $s_i$ and contextual observables $X_i$. This guides $g_W(X)$ to accurately fit the spacing distributions conditioned on various interaction contexts.
\begin{equation}\label{eq: log likelihood}
\begin{aligned}
    \mathcal{L}_{\mathrm{NLL}}&=-\ln\prod_{i=1}^N f(s_i;g_W(X_i))=\frac{1}{N}\sum_{i=1}^N\ell_{\mathrm{NLL}}(s_i,X_i),\\
    \ell_{\mathrm{NLL}}(s_i,X_i)&=\frac{1}{2}\left[\ln 2\pi+\ln \hat{\sigma}^2(X_i) + \frac{(\ln s_i - \hat{\mu}(X_i))^2}{\hat{\sigma}^2(X_i)}\right]+\ln s_i
\end{aligned}
\end{equation}

To avoid abrupt changes in risk quantification for continuous interactions, we introduce a smoothness regularisation term that penalises sharp differences in the distribution approximations for similar interaction contexts. Specifically, for each $X_i$, we generate a perturbed $X_i^\prime$ by adding a small Gaussian noise to each continuous variable in the original $X_i$. As shown in equation~\eqref{eq: smooth log likelihood}, then a Jensen-Shannon divergence ($D_\mathrm{JS}$) between the estimated distributions in context $X$ and a similar context $X^\prime$ is weighted by $\beta=5$ and added to $\mathcal{L}_\mathrm{NLL}$. Thereby, $W=\arg\min_W\mathcal{L}_\mathrm{SmoothNLL}$.
\begin{equation}\label{eq: smooth log likelihood}
\begin{aligned}
    \mathcal{L}_{\mathrm{SmoothNLL}}&=\frac{1}{N}\sum_{i=1}^N\ell_{\mathrm{SmoothNLL}}(s_i,X_i,X^\prime_i,\beta),\\
    \ell_{\mathrm{SmoothNLL}}(s_i,X_i,X^\prime_i,\beta)&=\ell_{\mathrm{NLL}}(s_i,X_i)+\beta D_\mathrm{JS}[f_S(s;g_W(X_i))||f_S(s;g_W(X_i^\prime))]\\
\end{aligned}
\end{equation}

\subsection{Context representation learning}\label{sec: context rl}
In this paper, we categorise the contextual observables $X$ describing an interaction scenario into three feature groups. \textbf{Current features} (denoted by $X_C$) represent the instantaneous states of interacting road users, including, e.g., their individual speeds and relative velocity. \textbf{Environment features} ($X_E$) describe external conditions during the interaction, such as weather, lighting, and road surface quality. \textbf{Historical kinematic features} ($X_T$) include time-series data of the interacting road users' speeds and yaw rates within the past 2.5 seconds. To avoid leakage of spacing information as suggested in~\cite{jiao2024unified}, random values in $X_T$ are dropped out (set as zero). A comprehensive list of these contextual features is provided in Appendix Table~\ref{tab: context features}. 

Each feature group is processed by a dedicated encoder, yielding encoded representation groups $\theta_C=g^C_{W_C}(X_C)$, $\theta_E=g^E_{W_E}(X_E)$, and $\theta_T=g^T_{W_T}(X_T)$. These groups are then concatenated and passed to a decoder $g^D_{W_D}$, which estimates conditional parameters as illustrated in Equation~\eqref{eq: decoder}. Details of these modules' architecture are provided in Appendix Section~\ref{sec: nn architecture}. 
\begin{equation}\label{eq: decoder}
\begin{aligned}
    \phi(X)&=g^D_{W_D}([\theta_C;\theta_E;\theta_T])
\end{aligned}
\end{equation}

The learning of encoded representation is implicitly achieved when training a $g_W(X)$ as a whole. In practice, the encoders can also be pretrained by separate representation learning. For instance, contrastive learning is well aligned with GSSM training, which essentially aims to capture spacing patterns in similar conditions and distinguish those in dissimilar conditions. To focus on presenting GSSM, we do not perform separate representation learning in the experiments in this paper, but future research is expected for more practical applications.

\subsection{Attribution of feature importance}\label{sec: attribution}
In order to improve traffic safety, the features that strongly influence the estimated risk of potential collisions are worth identifying. GSSM learns interaction patterns from real-world data and varies its estimates of collision risk in different contexts. To analyse these learnt patterns and attribute each feature's effect on the estimated risk, we use an explainable artificial intelligence (XAI) method known as Expected Gradients (EG~\cite{erion2021EG}).

EG extends Integrated Gradients~\cite{sundararajan2017IG} by taking the expectation of gradients over integration path steps $\alpha\sim U(0,1)$ and a set of reference points. For clearer interpretation, we compute attributions with respect to various pieces of latent representation $\theta$ (rather than raw inputs). The reference points are typically sampled from a uniform distribution over the entire training set~\cite{erion2021EG}. Here, we apply $k$-means clustering to $\Theta$, the representations encoded from all samples in the training set, and use cluster centres as representative references $\theta^\prime$. Equation~\eqref{eq: EG} summarises our computation of EG. The computed attributions sum to the difference in estimated risk given $\theta$ relative to the averaged risk given the references. The positive attribution of a feature means it contributes to increasing the estimated risk, whereas the negative attribution implies a contribution to decreasing the risk. More details about the encoder design to ensure correct dependence of contextual features are referred to Appendix Section~\ref{sec: nn architecture}.
\begin{equation}\label{eq: EG}
    \mathrm{EG}(\theta)=\mathbb{E}_{\theta^\prime\sim \Theta, \alpha\sim U(0,1)}\left\{(\theta-\theta^\prime)\frac{\partial}{\partial \theta}\log_{10}\left[\frac{\ln0.5}{\ln(1-F_S(s;g^D_{W_D}(\theta^\prime+\alpha(\theta-\theta^\prime)))}\right]\right\}
\end{equation}

Note that these attributions do not establish causality. For example, a positive attribution of relative speed does not imply that a lower relative speed would have reduced the risk. In a posteriori manner, EG explains the features' computational influence on the inference of a trained neural network. We recommend two strategies for interpretation. First, compare the attributions of all features at a specific time moment to assess their relative importance. Second, track how the attributions evolve over time. For instance, if the estimated collision risk rises along with an increasing attribution for a particular feature, that feature may have a relatively strong correlation with collision risk. Nonetheless, such an explanation alone cannot serve as evidence to justify an intervention, where future research is needed.

\section{Data of potential collisions}\label{sec: data}
Despite that GSSM does not use real-world traffic accidents for training purposes, a relatively small-scale dataset of crashes and near-crashes is necessary for effectiveness assessment. Such a dataset needs to contain both safe and safety-critical road user interactions in various situations. In Table~\ref{tab: crash datasets}, we summarise existing datasets that each have over 500 crashes and/or near-crashes recorded.
\begin{table}[htb]
\begin{minipage}{\linewidth}
\renewcommand{\thempfootnote}{\alph{mpfootnote}}
\renewcommand*\footnoterule{}
\centering
\caption{Existing datasets of crashes (and near-crashes).}
\label{tab: crash datasets}
\begin{tabular}{@{}llllll@{}}
\toprule
\multicolumn{1}{c}{\multirow{2}{*}{\textbf{Dataset} \footnote{Data sources: 
DAD, A3D, CCD, and DoTA were compiled from dashcam recordings posted on YouTube. 
DADA was compiled from dashcam recordings posted on YouTube, Youku, Bilibili, iQiyi, Tencent, etc. 
CADP was compiled from surveillance camera recordings posted on YouTube.}}} & \multicolumn{1}{c}{\multirow{2}{*}{\textbf{Year} \footnote{For SHRP2 NDS, the listed years are actual data collection time. 
For the other datasets, the years are publication time, since these sets were sourced from the Internet and have no determinate collection time.}}} & \multicolumn{1}{c}{\multirow{2}{*}{\textbf{Annotation} \footnote{Annotated information includes: event time (T), bounding box positions (S), crash type (C), participants (P), weather (W), lighting (L), driver's attention map (A). The datasets derived from SHRP2 NDS include additional information such as traffic density and road surface quality.}}} & \multicolumn{3}{c}{\textbf{Crashes (and near-crashes)}} \\ \cmidrule(l){4-6} 
\multicolumn{1}{c}{} & \multicolumn{1}{c}{} & \multicolumn{1}{c}{} & \multicolumn{1}{c}{\textbf{Number}} & \multicolumn{1}{c}{\textbf{Time range}} & \multicolumn{1}{c}{\textbf{Period before impact \footnote{Impact means physical contact in crashes; for near-crashes, the moment of closest proximity.}}} \\ \midrule
DAD~\cite{chan2016DAD} & 2016 & T & 620 & 5 s & Designed to be 4.5 s \\
CADP~\cite{shah2018CADP} & 2018 & T, S & 1,416 \footnote{Among these videos, 205 are annotated while the others are not.} & Varying & On average 3.7 s \\
A3D~\cite{yao2019A3D} & 2019 & T, P & 1,500 & Varying & Unrevealed \\
CCD~\cite{Bao2020CCD} & 2020 & T, P, W & 1,500 & 5 s & At least 3.0 s \\
DADA~\cite{Fang2019DADA-2000, Fang2022DADA} & 2019, 2022 & T, S, C, P, W, L, A & 2,000 & Varying & On average 5.2 s \\
DoTA~\cite{yao2022DoTA} & 2022 & T, S, C, P & 4,677 & Varying & On average around 4.0 s \\ \cmidrule{1-6}
SHRP2 NDS~\cite{Sears2019Honda,Layman2019DAS} & 2010-2013 & T, S, C, P, W, L, etc. & 8,895 \footnote{These include 1,942 crashes and 6,953 near-crashes.} & Varying & At least 20.0 s \\
Reconstructed SHRP2 & This paper & T, S, C, P, W, L, etc. & 6,664 \footnote{These include 1,402 crashes and 5,262 near-crashes.} & Varying & On average 22.8 s \\
\bottomrule
\end{tabular}
\end{minipage}
\end{table}

We select the database derived from the Second Strategic Highway Research Program’s (SHRP2) Naturalistic Driving Study (NDS), where we reconstruct bird's eye view trajectories for easier use. The major shortcoming of the other datasets is the lack of safe baselines to test false positives, except for DAD providing 1,130 clips sampled from source videos. Another issue is that most of the datasets were originally designed for accident detection rather than risk quantification, thus the time periods before crashes are not well preserved for assessing risk evolution. In the rest of this section, we first introduce SHRP2 NDS, then describe the trajectory reconstruction, and finally summarise the data we use in the following experiments. 

\subsection{SHRP2 Naturalistic Driving Study}
The SHRP2 NDS is a large-scale research initiative aimed at understanding driver behaviour and performance. From 2010 to 2013, it collected extensive data using instrumented vehicles in six states in the United States. More than 3,300 participant vehicles were equipped with data acquisition systems that recorded video footage, vehicle network data (e.g., speed, brake, and accelerator positions), and signals from additional sensors such as forward radar and accelerometers. A significant strength of SHRP2 NDS is its comprehensive set of manually annotated traffic safety events, including crashes and near-crashes that are collectively termed ``safety-critical events'', along with ``safe baselines'' selected through stratified random sampling. Table~\ref{tab: sce operational definition} outlines the operational definitions of these events as described in~\cite{SHRP2016}, where further details of the SHRP2 NDS are also referred to.
\begin{table}[htb]
\centering
\caption{Operational definitions of traffic safety events in the SHRP2 NDS.}
\label{tab: sce operational definition}
\begin{tabular}{@{}p{0.25\linewidth}p{0.72\linewidth}@{}}
\toprule
\textbf{Event} & \textbf{Operational definition} \\ \midrule
Crash & ``Any contact that the subject vehicle has with an object, either moving or fixed, at any speed in which kinetic energy is measurably transferred or dissipated is considered a crash. This also includes non-premeditated departures of the roadway where at least one tire leaves the paved or intended travel surface of the road, as well as instances where the subject vehicle strikes another vehicle, roadside barrier, pedestrian, cyclist, animal, or object on or off the roadway.'' \\ \cmidrule{1-2}
Near-crash & ``Any circumstance that requires a rapid evasive manoeuvrer by the subject vehicle, or any other vehicle, pedestrian, cyclist, or animal, to avoid a crash is considered a near-crash. A rapid evasive manoeuvrer is defined as steering, braking, accelerating, or any combination of control inputs.'' \\ \cmidrule{1-2}
Safe baseline (Non-conflict)& ``Normal driving behaviours and scenarios where the driver may react to situational conditions and events, but the reaction is not evasive and the situation does not place the subject or others at elevated risk.''\\ \bottomrule
\end{tabular}
\end{table}

\subsection{Trajectory reconstruction}\label{sec: SHRP2 reconstruction}
The accessible motion data from the SHRP2 NDS does not contain positional information of the subject (participant) vehicles to protect driver privacy. In addition, the other road users are detected by forward radars as surrounding objects in a subject-centric moving coordinate system. To align these data with the mainstream bird’s-eye-view datasets, we reconstruct the trajectories for both subject vehicles and surrounding objects using extended Kalman filters  (EKFs). First, we linearly interpolate all time-series signals to a uniform frequency of 0.1 seconds. Next, we reassign the indices of surrounding objects when necessary. Due to detection limitations, an object may be temporarily lost and re-detected later with a new index. If a newly detected object is within a specified distance threshold of a previously tracked object, the new object is assigned the previous index. We define the distance threshold as the object's position displacement relative to the subject over 0.3 seconds, constrained to a minimum of 0.5 m and a maximum of 2.5 m. Then we reconstruct trajectories for each event in two steps.

\textbf{Step 1: subject vehicle's trajectory reconstruction.} We apply an EKF assuming constant yaw rate and acceleration. The motion dynamics are updated according to equations~\eqref{eq: ekf_subject} and~\eqref{eq: ekf_subject_small_omega}, where $(x_i, y_i)$ denotes the vehicle’s position, $\psi_i$ the heading angle relative to the x-axis, $v_i$ the longitudinal speed, $\omega_i$ the yaw rate, $a_i$ the longitudinal acceleration, and $\epsilon=0.001$ is a threshold to use longitudinal updates when near-zero yaw rates induce numerical instability. The update interval $\Delta t = 0.1$ seconds. We place the subject vehicle initially at $(x_i,y_i)=(0,0)$ with its heading $\psi_i=0$ along the x-axis, and set the initial states of speed, yaw rate, and acceleration from original data. The data of yaw rates and accelerations were stably recorded, but speed measurements were not always consistent. We thus consider two orders of time sequence depending on whether the earliest or latest 0.5-second speed states are missing. If the latest states are missing, we let the EKF propagate forward from the earliest available measurement; if the earliest states are missing, we let the EKF propagate backwards. When both earliest and latest speed states exist, we run two EKFs from both ends and then select the reconstructed trajectory that deviates less in speed and yaw rate from the original sensor data.
{\small
\begin{equation}\label{eq: ekf_subject}
    \left[\begin{matrix}x_i\\ y_i\\ \psi_i\\ v_i\\ \omega_i\\ a_i
    \end{matrix}\right]_{t+\Delta t}=\left[\begin{matrix}x_i\\ y_i\\ \psi_i\\ v_i\\ \omega_i\\ a_i
    \end{matrix}\right]_{t}+
    \left[\begin{matrix}
    \Delta x_i\\
    \Delta y_i\\
    \omega_i\Delta t\\
    a_i\Delta t\\
    0\\
    0
    \end{matrix}\right]\text{, where}
\end{equation}}

{\small
\begin{equation}\label{eq: ekf_subject_small_omega}
\begin{aligned}
    \Delta x_i &= 
    \begin{cases}
    \begin{aligned}
        &\cos(\psi_i)(v_i\Delta t+\frac{1}{2}a_i\Delta t^2),&\text{if }|\omega_i|\leq\epsilon, \\
        &\frac{v_i\omega_i[\sin(\psi_i+\omega_i\Delta t)-\sin(\psi_i)] + a_i[\cos(\psi_i+\omega_i\Delta t)-\cos(\psi_i)] + a_i\omega_i\sin(\psi_i+\omega_i\Delta t)\Delta t}{\omega_i^2},&\text{otherwise};
    \end{aligned}
    \end{cases} \\
    \Delta y_i &= 
    \begin{cases}
    \begin{aligned}
    &\sin(\psi_i)(v_i\Delta t+\frac{1}{2}a_i\Delta t^2),&\text{if }|\omega_i|\leq\epsilon, \\
    &\frac{v_i\omega_i[\cos(\psi_i)-\cos(\psi_i+\omega_i\Delta t)] + a_i[\sin(\psi_i+\omega_i\Delta t)-\sin(\psi_i)] - a_i\omega_i\cos(\psi_i+\omega_i\Delta t)\Delta t}{\omega_i^2},&\text{otherwise}.
    \end{aligned}
    \end{cases} \\
\end{aligned}
\end{equation}}

\textbf{Step 2: surrounding objects’ trajectory reconstruction}. We then reconstruct the trajectories of the surrounding objects. This reconstruction has the forward field of view only, as only forward radar data are available. During an event, the subject vehicle may detect multiple objects (e.g., vehicles, cyclists, pedestrians, or animals), of which the edge nearest to the subject vehicle is detected. For each object, we first convert its local radar coordinates into the subject vehicle’s reconstructed global coordinate system. Then we determine the object's centroid based on its dimensions and whether the detected edge corresponds to its front or rear, inferred from its heading direction. Since only relative positions and speeds are available for these detected objects, we use an EKF under the assumption of constant heading and speed. The update equations are given in equation~\eqref{eq: ekf_sur}.
{\small
\begin{equation}\label{eq: ekf_sur}
    \left[\begin{matrix} x_j\\ y_j\\ \psi_j\\ v_j \end{matrix}\right]_{t+\Delta t} = \left[\begin{matrix} x_j\\ y_j\\ \psi_j\\ v_j \end{matrix}\right]_{t} +
    \left[\begin{matrix} \cos(\psi_j)v_j\Delta t \\ \sin(\psi_j)v_j\Delta t \\ \psi_j\\ v_j \end{matrix}\right]
\end{equation}}

Due to limited access to the raw data of cameras and radars, from which the provided data in the SHRP2 NDS were extracted, our reconstruction intends to preserve the first-hand information (i.e., the provided data) as much as possible. Therefore, this study defines reconstruction errors as the deviations of reconstructed trajectories from the provided signals in the SHRP2 NDS. We optimise the EKF parameters of uncertainties and motion ranges by minimising the root mean squared reconstruction error in subject speed, subject yaw rate, subject acceleration, and object speed, as well as the mean displacement error of objects. The distributions of eventual reconstruction errors are presented in Appendix Figure~\ref{fig: SHRP2_error_distributions}. Note that due to the absence of raw data, the very small errors we achieved do not indicate an accurate capture of reality, but of the original measurement. On average, the standard deviation is smallest for safe baselines and larger for near-crashes and crashes. But the difference is rather small, with 0.03-0.04 m/s in subject speed, 0.08-0.13 m/s$^2$ in subject acceleration, and 0.02-0.06 m/s in object speed.

\subsection{Data used in this paper}
We use crashes and near-crashes derived from the SHRP2 NDS as our test set for GSSM demonstration. The safe baselines are used as part of our training data. Therefore, this paper utilises all reconstructed event trajectories, alongside additional information relevant to the safety-critical events.

\begin{figure}[htb]
    \centering
    \includegraphics[width=\linewidth]{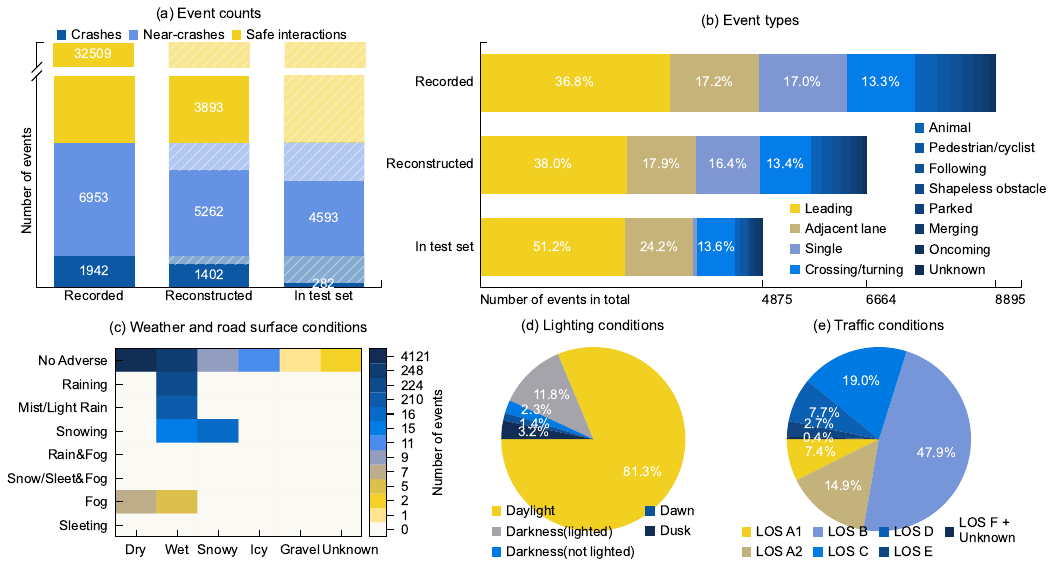}
    \caption{Statistics of original and processed events in the SHRP2 NDS. \textbf{(a)}~Numbers of crashes, near-crashes, and safe baselines that are recorded, reconstructed, and used in the test. \textbf{(b)}~Distribution of event types that are recorded, reconstructed, and used in the test. \textbf{(c)}~Distribution of weather and road surface conditions in the test events. \textbf{(d)}~Distribution of lighting conditions in the test events. \textbf{(e)}~Distribution of traffic conditions in the test events. LOS stands for level of service, details of which are referred to Appendix Table~\ref{tab: context features}.}
    \label{fig: SHRP2_event_counts}
\end{figure}

Not all 8,895 events in the original dataset could be reconstructed due to missing values, sensor inaccuracies, and the absence of ground-truth data. There are 6,664 events where the trajectories of both the subject vehicle and at least one surrounding object are reconstructed. From these, we extract a useful test set by excluding invalid events that meet any of the following criteria:
\begin{itemize}[noitemsep,topsep=-\parskip]
    \item no object is detected for more than 5 seconds (too short to observe risk evolution),
    \item the crash or near-crash is with an object behind the subject vehicle (given the lack of rearward radar data),
    \item the crash or near-crash involves a ``shapeless'' obstacle (e.g., roadside pavement).
\end{itemize}
Eventually, we obtain 4,875 safety-critical events in the test set. Figure~\ref{fig: SHRP2_event_counts}(a) presents the numbers of originally recorded events in the SHRP2 NDS, the subset with both subject and object trajectories reconstructed, and the final filtered subset. 

The distributions of event types are further displayed in Figure~\ref{fig: SHRP2_event_counts}(b), where the difference is primarily a result of the fact that rear-end events are more likely to be continuously detected by the forward radar. In Figures~\ref{fig: SHRP2_event_counts}(c-e), we also show the numbers and percentages of events in different environmental conditions. More than 80\% of the events occurred without adverse conditions, i.e., dry roads, not precipitation, and in daylight. However, considering that adverse conditions are less common by nature, they still play a role in causing traffic accidents.

In addition to the reconstructed trajectories, we attach other information about the safety-critical events. To consider a broader range of factors leading to potential collisions, the environmental conditions, including weather, lighting, road surface, and traffic density, are incorporated. To evaluate whether and when a potential collision is successfully detected, the annotations of time when an event starts and ends, when the driver reacts (if applicable), and when the impact occurs (a physical contact for crashes or the closest proximity in near-crashes) are also included. Further, the specific type (e.g., with a leading object, during crossing or turning) of an event is included to evaluate the detection of potential collisions in different scenarios. Lastly, when necessary, each event's narrative is referred to for further verification. These result in the largest to date trajectory dataset of traffic crashes and near-crashes. This dataset has been made accessible in \cite{jiao2025shrpcrash}. 

\section{Experiments}\label{sec: exps}
We run all experiments with an NVIDIA A100 GPU (80GB RAM) as well as 5 to 50 Intel Xeon CPUs, depending on whether parallel computation is helpful. To ensure fair comparison across settings and methods, we control the following conditions: random seed, maximum training epochs, early stopping criteria, and evaluation sample sets. Hyperparameter choices are listed in Appendix Table~\ref{tab: hyperparameters}. We do not conduct an ablation experiment for neural network design, as our core focus in experiments is on validating the proposed GSSM.

\subsection{Training datasets and experiment design}
Our training data consist of 3 datasets collected in different countries, involving different interaction scenarios, and with different equipment. These data significantly differ from the test set, and we deliberately exploit this difference to demonstrate the generalisability of GSSM. In addition, the diversity of these data mimics the heterogeneity in real-world traffic interactions. 

The first dataset is composed of the safe baselines derived from SHRP2 NDS as described in Section~\ref{sec: data}. We term this set as \emph{SafeBaseline}, where the interactions were recorded both on highways and in urban traffic. Considering the detection limitation of forward radar, the majority of SafeBaseline may be on straight roads. To compensate for that, we introduce two more datasets. One is specifically focused on urban intersections. This dataset was collected in the U.S. in 2019 by a fleet of automated vehicles~\cite{argoverse}. We use the set of interactions between human road users extracted by~\cite{argo_int}, and term this set as \emph{ArgoverseHV}. From the other dataset known as \emph{highD}, we extract lane-change interactions using the same methods in our previous work~\cite{jiao2024unified}. This dataset was collected by drones on German highways in 2018, with a position error typically less than 10 cm~\cite{highd}. To make fair use, we draw approximately equal-sized (over 522 thousand) samples for each dataset, where 80\% are used for training and 20\% for model validation to monitor training progress.

The experiments are designed as follows to demonstrate the effectiveness of GSSM and its useful characteristics targeted in this research.
\begin{itemize}[noitemsep,topsep=-\parskip]
    \item \textbf{Effectiveness}. Effective risk quantification accurately identifies dangers and provides timely alerts to prevent potential collisions. Therefore, we compare GSSM with other existing methods on the \textit{accuracy} in alerting crashes and near-crashes as well as the \textit{timeliness} of alerts. For comparability, the GSSM compared in this experiment is trained using only current features ($X_C$).
    \item \textbf{Scalability}. The training of GSSM and its effectiveness are expected to be scaled with increased interaction patterns. We combine SafeBaseline with varying proportions (random sampling 10\%--100\%) of ArgoverseHV or highD, train a series of GSSMs on the combined datasets, and then observe the variation in their effectiveness.
    \item \textbf{Context-awareness}. GSSM is capable of considering varying contextual information, and additional contextual features are expected to enhance risk quantification effectiveness. We train multiple GSSMs on SafeBaseline, which uniquely provides environment data. In addition to current features ($X_C$), these GSSMs progressively incorporate more contextual information of subject acceleration ($a_{i}$), environment features ($X_E$), and historical kinematic features ($X_T$). 
    \item \textbf{Generalisability}. Our training of GSSM does not use any crashes or near-crashes as in the test set. Essentially, the effectiveness of all the GSSMs trained in this study is generalised from normal interactions, which vary in location and equipment during data collection. In addition to that, GSSM is designed to be generally applicable across interaction scenarios. We compare the effectiveness of GSSM against other methods for different types of safety-critical events, including rear-end scenarios, lateral interactions with other road users in adjacent lanes, crossing and turning conflicts, merging situations, and a small number of incidents involving vulnerable road users (e.g., pedestrians and cyclists).
    \item \textbf{Risk attribution}. Lastly, we evaluate the dominant contributing factors to the change in collision risk based on feature attribution.
\end{itemize}
\vspace{\parskip}

Table~\ref{tab: gssms2learn} summarises the 29 GSSMs that are trained across these experiments. These variations depend on the training datasets (or their combinations) and the selected contextual observables.
\begin{table}[htbp]
\begin{minipage}{\linewidth}
\renewcommand{\thempfootnote}{\alph{mpfootnote}}
\renewcommand*\footnoterule{}
\centering
\caption{Overview of GSSMs trained in the experiments and compared with existing methods.}
\label{tab: gssms2learn}
\resizebox{\textwidth}{!}{%
\begin{tabular}{l@{}l@{}l@{}c@{}c@{}c@{}c@{}}
\toprule
\multicolumn{1}{c}{\textbf{Acronym} \footnote{The two-dimensional surrogate safety measures (2D SSMs) have abbreviations TTC2D for Two-dimensional Time-to-Collision~\cite{guo2023ttc2d}, TAdv for Time Advantage~\cite{laureshyn2010tadv}, ACT for Anticipated Collision Time~\cite{venthuruthiyil2022act}, and EI for  Emergency Index~\cite{cheng2025ei}.}\footnote{Acronyms for GSSMs follow the format \texttt{<Datasets>-<Contextual observables>}. For example, ``S-C'' is trained on SafeBaseline and uses current features $X_C$}} & \multicolumn{1}{c}{\textbf{Training dataset(s)} \footnote{The dataset names are abbreviated as ``S'' = SafeBaseline, ``A'' = ArgoverseHV, ``h'' = highD. If multiple datasets are used (e.g., ``SAh''), the GSSM is trained on their union.}} & \multicolumn{1}{c}{\textbf{\begin{tabular}[c]{@{}c@{}}Contextual\\ observables\end{tabular}}} & \multicolumn{1}{c}{\textbf{\begin{tabular}[c]{@{}c@{}}Experiment(s)\\ involved in\end{tabular}} \footnote{The experiments are abbreviated as ``E''=Effectiveness, ``S''=Scalability, ``C''=Context-awareness, ``G''=Generalisability, ``R''=Risk attribution. A GSSM may be involved in multiple experiments.}} & \multicolumn{1}{c}{\textbf{\begin{tabular}[c]{@{}c@{}}Number of\\ parameters\end{tabular}}} & \multicolumn{1}{c}{\textbf{\begin{tabular}[c]{@{}c@{}}Training time\\ per sample (s)\end{tabular}}} & \multicolumn{1}{c}{\textbf{\begin{tabular}[c]{@{}c@{}}Inference time\\ per sample (s)\end{tabular}}} \\ 
\midrule
TTC2D & \multirow{4}{*}{N/A} & \multirow{10}{*}{$[X_C]$} & \multirow{4}{*}{E, G} & \multirow{4}{*}{N/A} & \multirow{4}{*}{N/A} & 6.21E-7 \\
ACT &  &  & &  &  & 3.93E-6 \\
TAdv &  &  & &  &  & 4.78E-6 \\
EI &  &  & &  &  & 5.05E-4 \\
S-C & SafeBaseline &  & S, C & 773,286 & 4.43E-5 & 1.37E-5 \\
h-C & highD &  & E, G & 773,286 & 4.39E-5 & 1.07E-5 \\
A-C & ArgoverseHV &  & G & 773,286 & 4.40E-5 & 1.34E-5 \\
SA-C & SafeBaseline $\cup$ ArgoverseHV$^{\{10\%,20\%,\cdots,100\%\}}$\footnote{The data combined in addition to SafeBaseline are randomly sampled for 10\%-100\% from the whole dataset.} &  & S & 773,286 & \multirow{2}{*}{Not measured} & \multirow{2}{*}{Not measured} \\
Sh-C & SafeBaseline $\cup$ highD$^{\{10\%,20\%,\cdots,100\%\}}$ &  & S & 773,286 &  &  \\
SAh-C & SafeBaseline $\cup$ 10\%ArgoverseHV $\cup$ 100\%highD &  & S & 773,286 & 4.53E-5 & 1.19E-5 \\
S-Ca & \multirow{5}{*}{SafeBaseline} & $[X_C, a_i]$ & C & 775,336 & 5.04E-5 & 1.33E-5 \\
S-CE &  & $[X_C, X_E]$ & C & 857,486 & 6.42E-5 & 1.78E-5 \\
S-CaE &  & $[X_C, a_i, X_E]$ & C & 859,536 & 6.62E-5 & 1.88E-5 \\
S-CET &  & $[X_C, X_E, X_T]$ & C, R, G & 885,664 & 7.48E-5 & 2.21E-5 \\
S-CaET &  & $[X_C, a_i, X_E, X_T]$ & C & 887,714 & 7.50E-5 & 2.35E-5 \\ 
\bottomrule
\end{tabular}}
\end{minipage}
\end{table}

Since we do not include video data yet as contextual information, the comparison in this paper's experiments excludes the self-supervised methods based on computer vision or anomaly detection. Instead, we compare with surrogate safety measures (SSMs) tailored for two-dimensional (longitudinal and lateral) interactions as listed in Table~\ref{tab: 2dssms}. Their estimations are all based on the instantaneous states of interacting road users, i.e., the current features $X_C$. For efficient large-scale evaluation, we implement TAdv, ACT, and TTC2D in a vectorised form, enabling parallel computation over thousands of interaction pairs.
\begin{table}[htbp]
\centering
\caption{Two-dimensional Surrogate Safety Measures (2D SSMs) used as baseline methods.}
\label{tab: 2dssms}
\begin{tabular}{@{}llp{0.57\linewidth}@{}}
\toprule
\multicolumn{1}{c}{\textbf{Two-dimensional SSM}} & \multicolumn{1}{c}{\textbf{Year}} & \multicolumn{1}{c}{\textbf{Brief summary}} \\ \midrule
Time Advantage (TAdv)~\cite{laureshyn2010tadv} & 2010 & Expected time gap between two road users passing a common spatial zone, assuming no behaviour change in the short future. This is also known as predicted Post-Encroachment-Time (PET). \\ \midrule
Anticipated Collision Time (ACT)~\cite{venthuruthiyil2022act} & 2022 & The shortest distance between two road users divided by their closing-in rate, assuming a constant closing-in rate in the short future. \\ \cmidrule{1-3}
\begin{tabular}[t]{@{}l@{}}Two-dimensional Time-to-Collision \\ (TTC2D)~\cite{guo2023ttc2d}\end{tabular} & 2023 & Minimum of the longitudinal and lateral Time-to-Collision values, assuming no behaviour change in the short future. \\ \cmidrule{1-3}
Emergency Index (EI)~\cite{cheng2025ei} & 2025 & Intensity of evasive action required to prevent a potential collision, based on the change rate in overlapping path. \\ \bottomrule
\end{tabular}
\end{table}

There also exist extensions that incorporate stochastic behaviour and/or extreme value theory (e.g., survival analysis), such as~\cite{eggert2017,puphal2019,deGelder2023}. We do not compare with these methods because their implementations require specific modelling assumptions about uncertainty (for example, how to represent trajectory noise and behavioural variability), which makes a fair and consistent comparison difficult within our current framework. Nevertheless, we refer interested readers to this line of work as a complementary direction that combines SSMs and probabilistic modelling.

\subsection{Ground truth of safety-critical events}
\begin{figure}[htb]
    \centering
    \includegraphics[width=6.5in]{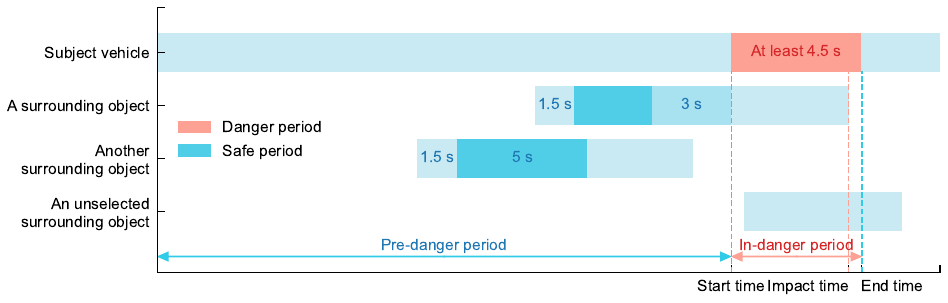}
    \caption{Separation of time periods for each safety-critical event in the test set for evaluation.}
    \label{fig: Safe_danger_period}
\end{figure}

The original SHRP2 NDS annotates the time moments of the start, impact, and end of each crash or near-crash. For crashes, impact time means physical contact time; for near-crashes, impact time is when the road users were at the closest proximity. The safety-critical events in the test set have at least 20 seconds before the impact time. This allows for separating the danger period and safe period in each event. Then we can create positive ground truth (i.e., potential collision) between the subject vehicle and a conflicting object in the danger period, and negative ground truth (i.e., safe interactions) between the subject vehicle and other surrounding objects in the safe period. As illustrated in Figure~\ref{fig: Safe_danger_period}, we consider the danger period of an event from the start time, or 4.5 seconds before impact time if the annotated start time is less than 4.5 seconds until impact. This danger period is considered to finish at the end time, or in 0.5 seconds after impact time if the annotated end time is delayed. Meanwhile, we consider the safe period for each surrounding object as 2 to 5 seconds after the object is detected for 1.5 seconds and 3 seconds before the event start time. To be considered as in a safe period, the object should not have a hard braking (deceleration greater than 1.5 m/s$^2$).

The specific ``conflicting object'' in an event is not explicitly annotated in the SHRP2 data, and it's possible the conflicting object was not detected or recorded. We design a three-stage evaluation procedure to filter out those invalid cases, create useful ground truth, and then perform a reliable evaluation. 
\begin{itemize}[noitemsep,topsep=-\parskip]
    \item \textbf{In the first stage}, for each event, we use every method among the 2D SSMs and GSSMs (except for the ones used to test scalability only) to evaluate the collision risk between the subject vehicle and each of all detected surrounding objects, then determine a temporary conflicting object based on the evaluation. An event may have zero or a few candidate objects evaluated by different methods. 
    \item \textbf{In the second stage}, we let the methods vote for the most probable conflicting object. Considering that each method votes for a candidate object, an abstention occurs when a method determines that no temporary conflicting object exists. We select the object earning the most votes to be the eventual conflicting object. This voted object should earn over 1/3 of the total votes, and have less than 1/3 of the total votes against. 
    \item \textbf{In the third stage}, we apply the 2D SSMs and GSSMs to each event that has a voted conflicting object. True and false positives are assessed based on the potential collision between the subject vehicle and the voted conflicting object in the danger period. True and false negatives are assessed based on the safe interactions between the subject vehicle and other surrounding objects in their safe periods.
\end{itemize}
\vspace{\parskip}

To determine a temporary conflicting object in the first stage, we assume that the conflicting object in an event is the most risky object in the danger period, and its collision risk is less in the pre-danger period than in-danger period. More specifically, the object with the highest average collision risk during in-danger period is first selected. If the selected object has no data before the danger period, further confirmation is not possible, so we skip the event to be conservative than misleading. If any data of this selected object is recorded before danger, the 25th, 50th, and 75th percentiles of the risk quantification in the pre-danger period should be less than those in the danger period; otherwise, this object is not the real conflicting object. Here we use the percentiles (rather than, e.g., minimum, mean, maximum) for a more robust comparison and to avoid any potential influence of outliers. 

\subsection{Evaluation metrics}
As mentioned in Section~\ref{sec: problem}, the evaluation of proactive risk quantification includes two aspects of detection accuracy and alert timeliness. The following subsections detail the metrics used in this paper.

\subsubsection{Detection accuracy}
Using a threshold, we convert the risk quantification of a potential collision into binary outcomes of ``safe'' or ``unsafe'', then use the unsafe outcomes to issue alerts. Depending on whether the alerts correctly warn of a crash or near-crash and stay silent during safe interactions, each event has 4 possible counts. Targeting the conflicting object in the danger period, a True Positive (TP) issues alerts for at least 0.5 seconds; a False Negative (FN) issues no alerts. For all qualified objects other than the conflicting object in their safe periods, True negatives (TNs) issue no alerts; False positives (FPs) incorrectly issue alerts. Then counting for all test events, equation~\eqref{eq: fnr and fpr} defines false negative rate $R_\mathrm{FN}$ and false positive rate $R_\mathrm{FP}$; equation~\eqref{eq: pre rec and f1} defines $\mathrm{Precision}$, $\mathrm{Recall}$, and a combined accuracy score $F_1$. 
\begin{equation}\label{eq: fnr and fpr}
    R_\mathrm{FN} = \frac{\mathrm{FN}}{\mathrm{TP}+\mathrm{FN}},\quad R_\mathrm{FP} = \frac{\mathrm{FPs}}{\mathrm{FPs}+\mathrm{TNs}}.
\end{equation}
\begin{equation}\label{eq: pre rec and f1}
    \mathrm{Precision}=\frac{\mathrm{TP}}{\mathrm{TP}+\mathrm{FPs}},\quad \mathrm{Recall} = \frac{\mathrm{TP}}{\mathrm{TP}+\mathrm{FN}},\quad F_1=\frac{2\mathrm{Precision}\cdot\mathrm{Recall}}{\mathrm{Precision}+\mathrm{Recall}}.
\end{equation}

\subsubsection{Alert timeliness}
We define Time to Impact (TTI) to measure how early a potential collision is recognised relative to impact time ($t_\mathrm{impact}$), the moment of physical contact in crashes or the moment of minimum proximity in near-crashes. As a posteriori quantity, TTI counts from the last time moment when risk quantification shifts from safe to unsafe until a (near-)crash in reality. This is formally defined in equation~\eqref{eq: tti}, where $M_t$ is the estimated risk level of a potential collision at time $t$, $M^*$ is a threshold to distinguish safety. We let $M_{-1}\leq M^*$ in case all recorded time moments are estimated as unsafe. Without loss of generality, we assume that a larger $M$ indicates a higher risk; the inequalities can be reversed if a smaller $M$ indicates higher risk.
\begin{equation}\label{eq: tti}
    \mathrm{TTI}=t_\mathrm{impact}-\max\{t\mid M_t>M^*, M_{t-1}\leq M^*,t\leq t_\mathrm{impact}\}
\end{equation}

For a collective evaluation over test events, we first use $P_{\mathrm{TTI}\geq1.5}$, which indicates the percentage of events with $\mathrm{TTI}\geq1.5$ s among correctly detected (near-)crashes. The threshold of 1.5 s is motivated by empirical findings that human drivers typically require 1 to 1.3 s to respond to an obstacle \cite{Summala2000,Markkula2016}. As a second measure of alert timeliness, we use median time-to-impact across all correctly detected (near-)crashes, denoted by $m\mathrm{TTI}$. Based on the definition in equation~\eqref{eq: tti}, $\mathrm{TTI}\geq0$ and its maximal value depends on the time period recorded before impact. We thus restrain $\mathrm{TTI}<10$ s when calculating $m\mathrm{TTI}$. To reflect the variation of $\mathrm{TTI}$, we report $m\mathrm{TTI}$ $[Q1, Q3]$; $99\%CI$, where $Q1$ and $Q3$ are the first and third quartiles, respectively, and $99\%CI$ is the 99\% confidence interval obtained via two-sided sign test for $m\mathrm{TTI}$. Similar to the accuracy metrics, $m\mathrm{TTI}$ and $P_{\mathrm{TTI}\geq1.5}$ vary at different alerting thresholds.

\subsubsection{Performance curves}
Risk quantification must balance the trade-off between desired yet conflicting characteristics at different thresholds. For example, a threshold resulting in smaller $R_\mathrm{FP}$ necessarily increases $R_\mathrm{FN}$. Such a threshold also entails higher $\mathrm{Precision}$, but lower $\mathrm{Recall}$. Likewise, earlier alerts (i.e., larger $\mathrm{TTI}$) tend to be accompanied by more false positives and less accurate detection. Therefore, our evaluation uses three key performance curves to compare different methods on a common set of events in the test dataset. Each curve is obtained by varying the threshold that distinguishes risk from safety. As summarised in Table~\ref{tab: trade-off curves}. Receiver operating characteristic curve (ROC) intuitively presents the trade-off between false detection rates; Precision-recall curve (PRC) comprehensively evaluates accuracy performance; and Accuracy-timeliness curve (ATC) evaluates timeliness performance under different accuracy levels.
\begin{table}[htbp]
\centering
\caption{Performance curves of risk quantification at varying thresholds.}
\label{tab: trade-off curves}
\begin{tabular}{@{}llll@{}}
\toprule
\textbf{Performance curve} & \textbf{Horizontal axis} & \textbf{Vertical axis} & \textbf{Optimal point} \\ \midrule
Receiver operating characteristic curve (ROC) & $R_\mathrm{FP}\in[0,1]$ & $1-R_\mathrm{FN}=R_\mathrm{TP}\in[0,1]$ & In theory, $(0,1)$ \\
Precision-recall curve (PRC) & $\mathrm{Recall}\in[0,1]$ & $\mathrm{Precision}\in[0,1]$ & In theory, $(1,1)$ \\
Accuracy-timeliness curve (ATC) & $m\mathrm{TTI}\in[0,10]$ & $F_1\in[0,1]$ & $(\max F_1, m\mathrm{TTI}^*)$ \\ \bottomrule
\end{tabular}
\end{table}

If a perfect model exists, ROC and PRC have their theoretical optimal points of zero false and all correct detection. ATC does not have a theoretical optimum, but we consider a practical optimal point to be the $m\mathrm{TTI}$ when $F_1$ is the highest, defined as $m\mathrm{TTI}^*=\arg\max_{\mathrm{TTI}}F_1$. Accordingly, we denote the $P_{\mathrm{TTI}\geq1.5}$ at this point $P^*_{\mathrm{TTI}\geq1.5}$.

\subsubsection{Safety-focused evaluation metrics}
Based on the performance curves, we first use AUPRC, the area under PRC, as a comprehensive metric to assess detection accuracy. Here we do not consider the area under ROC, as a method that has a better-performed PRC (i.e., of larger AUPRC) necessarily has a larger area under ROC~\cite{davis2006prc}. For timeliness evaluation, we use $P^*_{\mathrm{TTI}\geq1.5}$ and $m\mathrm{TTI}^*$ when $F_1$ is at the highest point on ATC. The larger the values, the earlier the alerts under the optimal detection accuracy. 

Risk alerting of potential collisions is different from normal binary classification problems, as false negatives can cause significantly more severe consequences than false positives. To emphasise the high cost of false negatives, we define two safety-focused metrics based on ROC and PRC, respectively. The first is $A^\mathrm{ROC}_R$ as defined in equation~\eqref{eq: a_roc}, the area under the ROC curve when $1-R_\mathrm{FN}$ is larger than $R$ (i.e., false negative rate is smaller than $1-R$). For easier comparison, we normalise the area relative to its maximum possible value $1-R$. The second is $\mathrm{Precision}^\mathrm{PRC}_R$ defined in equation~\eqref{eq: r_prc}, the highest $\mathrm{Precision}$ on the PRC when $\mathrm{Recall}$ is larger than $R$. For both metrics, the larger the value, the more accurate a model is, given a necessarily low false negative rate. 
\begin{equation}\label{eq: a_roc}
    A^\mathrm{ROC}_{R}=\frac{1}{1-R}\int_R^1\left(1-R_\mathrm{FP}(r)\right)\mathrm{d}r,\text{ with }r=1-R_\mathrm{FN}
\end{equation}
\begin{equation}\label{eq: r_prc}
    \mathrm{Precision}^\mathrm{PRC}_R=\max(\mathrm{Precision}\mid\mathrm{Recall}\geq R)
\end{equation}

To offer an overview, we use $\mathrm{AUPRC}$, $A^\mathrm{ROC}_{80\%}$, $A^\mathrm{ROC}_{90\%}$, $\mathrm{Precision}^\mathrm{PRC}_{80\%}$, and $\mathrm{Precision}^\mathrm{PRC}_{90\%}$ to evaluate detection accuracy, while $P^*_{\mathrm{TTI}\geq1.5}$ and $m\mathrm{TTI}^*$ are used for timeliness evaluation. We select the restriction rate $R$ to be 80\% and 90\% to reinforce a focus on safety. Some methods, if they are not safe enough, can have zero area measured by $A^\mathrm{ROC}_{80\%}$ or $A^\mathrm{ROC}_{90\%}$, or are not applicable (N/A) when measured by $\mathrm{Precision}^\mathrm{PRC}_{80\%}$ or $\mathrm{Precision}^\mathrm{PRC}_{90\%}$. For all of the metrics, higher values indicate better performance in collision risk quantification.

\section{Results}\label{sec: results}
The risk quantification of safety-critical interactions by GSSM is generalised from normal traffic interactions. The training of GSSM is on normal interaction data, of which the collection varies in locations and equipment. Then the test set contains thousands of crashes and near-crashes where at least 20 seconds are recorded before a safety-critical event happens, offering both positive and negative ground truth. In the experiment on Scalability, we use the first-stage evaluation results of the 4,875 safety-critical events in the test set, and determine a proportion of 10\% to add ArgoverseHV data and 100\% for highD (see Section~\ref{sec: exp2} for the results), as already noted in Table~\ref{tab: gssms2learn}. Then the settlement of ground truth excludes GSSMs trained on other proportions of data combinations. Out of the 4,875 test events, 110 crashes and 2,481 near-crashes eventually have their conflicting objects determined. These 2,591 events form a reliable evaluation basis for the other experiments on Effectiveness (Section~\ref{sec: exp1}), Context-awareness (Section~\ref{sec: exp3}), Generalisability (Section~\ref{sec: exp4}), and Risk attribution (Section~\ref{sec: exp5}), where the third-stage evaluation results are presented. In the reliable set, each of the 2,591 events has a danger period as a positive sample, and 3,777 safe periods with different surrounding objects serve as negative samples. These events are considered distinct observations, and every method is applied to a single event only once.

\subsection{Effective risk quantification}\label{sec: exp1}
\begin{figure}[htbp]
    \centering
    \includegraphics[width=0.7\linewidth]{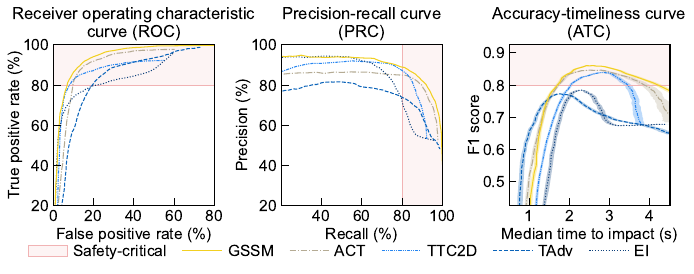}
    \caption{\textbf{Performance curves comparing GSSM and existing two-dimensional surrogate safety measures (2D SSMs)}. The GSSM under comparison is h-C which is trained on the highD data and uses current features. For the other 2D SSMs, TAdv is abbreviated for Time Advantage; ACT for Anticipated Collision Time; TTC2D for Two-dimensional Time-to-Collision; and EI for Emergency Index. Each method is tested on the same 2,591 safety-critical events, including 2,481 near-crashes and 110 crashes. In the ATC plot, the shaded bands represent 99\% confidence intervals for median time to impact. }
    \label{fig: Result1}
\end{figure}

We use Figure~\ref{fig: Result1} and Table~\ref{tab: Result1} to show the effectiveness of GSSM in detection accuracy and alert timeliness. Compared with other existing two dimensional surrogate safety measures (2D SSMs), GSSM achieves superior accuracy in detecting (near-)crashes, while maintaining relatively stable timeliness of alerts. As is shown by the ROC and PRC in Figure~\ref{fig: Result1}, the curves of GSSM enclose the curves of all the other methods. This means an absolutely higher accuracy, with fewer false positives and false negatives. Regarding alert timeliness, as is seen in Table~\ref{tab: Result1}, the percentage of early alerts issued 1.5 s in advance and the median time to impact at optimal accuracy (i.e., $P^*_\mathrm{TTI\geq1.5}$ and $m\mathrm{TTI}^*$) of GSSM are slightly lower than those of ACT and TTC2D. But GSSM exhibits a narrower interquartile range and 99\% confidence interval, indicating more stable performance across events. The accuracy-timeliness curves in Figure~\ref{fig: Result1} also show that GSSM consistently secures a longer time for collision prevention than the other methods at the same accuracy, as indicated by the $F_1$ score, higher than 0.8.
\begin{table}[htbp]
\caption{\textbf{Numeric performance evaluation of GSSM and existing methods}. Each method is tested on the same 2,591 safety-critical events, including 2,481 near-crashes and 110 crashes. The 99\% confidence intervals (CI) of $m\mathrm{TTI}^*$ are obtained via two-sided sign test. For all of the metrics, higher values indicate better performance. The best value in each column is underlined and marked bold; the second-best value is marked bold.}
\label{tab: Result1}
\centering
\begin{tabular}{@{}lcccccccc@{}}
\toprule
\textbf{Method} & AUPRC & $A_{80\%}^\mathrm{ROC}$ & $A_{90\%}^\mathrm{ROC}$ & $\mathrm{Precision}_{80\%}^\mathrm{PRC}$ & $\mathrm{Precision}_{90\%}^\mathrm{PRC}$ & $P^*_{\mathrm{TTI}\geq1.5}$ & $m\mathrm{TTI}^*$ $[Q1, Q3]$; $99\%CI$ \\
\midrule
GSSM & \textbf{\underline{0.900}} & \textbf{\underline{0.817}} & \textbf{\underline{0.729}} & \textbf{\underline{0.887}} & \textbf{\underline{0.814}} & \textbf{0.819} & 2.60 [1.80, 3.51]; 2.53--2.65 \\
ACT & 0.824 & \textbf{0.721} & \textbf{0.561} & 0.850 & \textbf{0.795} & 0.769 & \textbf{2.78} [1.63, 3.91]; 2.70--2.88 \\
TTC2D & 0.818 & 0.493 & 0.139 & \textbf{0.865} & 0.674 & \textbf{\underline{0.860}} & \textbf{\underline{2.83}} [1.97, 3.83]; 2.75--2.90 \\
TAdv & 0.700 & 0.584 & 0.414 & 0.738 & 0.640 & 0.606 & 1.73 [0.99, 2.50]; 1.68--1.78 \\
EI & \textbf{0.842} & 0.426 & 0.251 & 0.706 & 0.539 & 0.747 & 2.26 [1.47, 3.20]; 2.20--2.34 \\
\bottomrule
\end{tabular}
\end{table}

\subsection{Scalable GSSM training}\label{sec: exp2}
\begin{figure}[htbp]
    \centering
    \includegraphics[width=\linewidth]{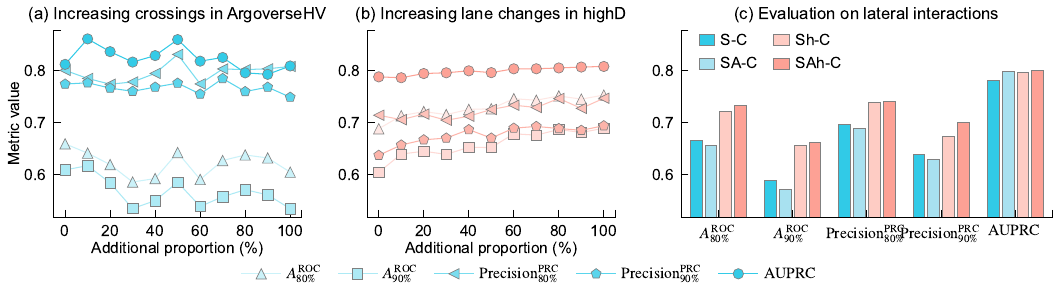}
    \captionof{figure}{Variation in risk quantification accuracy of GSSMs trained on different combinations of interaction patterns. \textbf{(a)}~Accuracy changes when the training data of SafeBaseline are combined with different proportions of ArgoverseHV data, which majorly cover crossing and turning interactions at urban intersections. The accuracy evaluations are performed for event types of Crossing/turning and Pedestrian/cyclist. \textbf{(b)}~Accuracy changes when the training data of SafeBaseline are combined with different proportions of highD data, which majorly cover lane-change interactions. The accuracy evaluations are performed for event types of Adjacent lane and Merging. \textbf{(c)}~Accuracy evaluations with events involving lateral interactions. All the GSSMs being compared use current features. The compared models include S-C that is trained on SafeBaseline only, SA-C that combines SafeBaseline and 10\% ArgoverseHV, Sh-C that combines SafeBaseline and 100\% highD, and SAh-C that combines SafeBaseline, 10\% ArgoverseHV, and 100\% highD.}
    \label{fig: Result2}
\end{figure}

Without requiring crashes or manual labels of near-crashes for training, GSSM leverages normal interactions, of which the available data is far more abundant. This not only means GSSM learns interaction patterns, but also implies effectiveness improvement by feeding more diverse patterns of traffic interactions. Figure~\ref{fig: Result2} illustrates such scalability of GSSM in accurately alerting potential collisions. The GSSMs being compared are all trained with instantaneous states of road user movement. We use the GSSM trained on the SafeBaseline dataset as a benchmark, and increasingly include more data from the crossing interactions in the ArgoverseHV dataset and lane-changes in the highD dataset. 

Figure~\ref{fig: Result2}(a) shows the variation in alerting accuracy as increasing interactions of crossing and turning in ArgoverseHV are included; while Figure~\ref{fig: Result2}(b) shows a counterpart for gradually including lane-change interactions in highD. A clear and consistent improvement is seen when more lane changes in highD serve training. As the proportion of ArgoverseHV data increases, however, the accuracy metrics are first enhanced and then drop. This is reasonable because crossing and turning interactions account for around 13.6\% in the test set (see Figure~\ref{fig: SHRP2_event_counts}). While these interactions are increasingly included, the training of GSSM gradually shifts its optimisation direction, which may undermine the patterns learnt from the SafeBaseline data.

Based on Figure~\ref{fig: Result2}(a) and \ref{fig: Result2}(b), we consider 10\% as the best proportion to include additional data from ArgoverseHV and 100\% from highD. Then Figure~\ref{fig: Result2}(c) compares the enhancement of risk quantification accuracy by including more training data in addition to SafeBaseline. It is clearly seen that, to different degrees, incorporating additional lateral interaction data improves the detection accuracy of crashes and near-crashes involving lateral interactions. This implies exciting future work to obtain more powerful GSSMs by training on larger-scale interaction data.

\subsection{Context-aware GSSM}\label{sec: exp3}
\begin{figure}[hbtp]
    \centering
    \includegraphics[width=\linewidth]{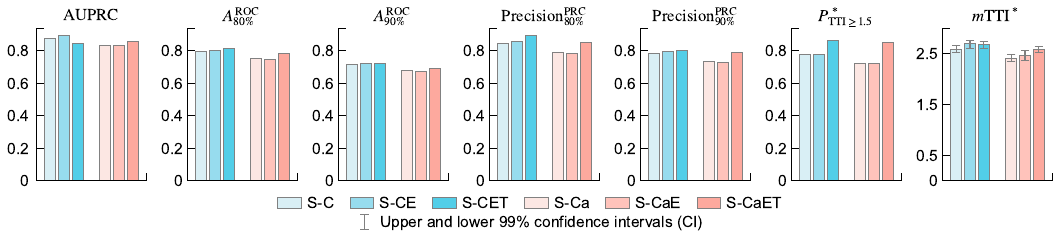}
    \captionof{figure}{Accuracy and timeliness comparison of GSSMs that consider varying contextual information. These GSSMs are all trained on SafeBaseline. S-C uses current features only; S-CE uses current and environment features; S-CET uses current, environment, and historical kinematic features. The other three, i.e., S-Ca, S-CaE, and S-CaET, additionally consider the instantaneous acceleration of the subject vehicle. The 99\% confidence intervals (CI) of $m\mathrm{TTI}^*$ are obtained via two-sided sign test.}
    \label{fig: Result3}
\end{figure}

Collision risk is conditioned on interaction context. For example, a usually safe interaction may become unsafe if it is raining. As another example, the same speed could be safe on highways while unsafe in urban traffic. Therefore, being aware of the context where an interaction happens is important, and GSSM allows for that. Figure~\ref{fig: Result3} compares all the evaluation metrics of alerting accuracy and timeliness across GSSMs that consider different contextual information. In each plot specific to a metric, we separate two groups of GSSMs. One group does not consider the acceleration of the subject vehicle as a current feature, while the other group does. Within each group, we progressively include additional features of environmental conditions and historical kinematics.

The GSSMs using only instantaneous motion kinematics serve as baselines for comparisons within each group. Additionally using environmental conditions in general achieves better performance in terms of both detection accuracy and alert timeliness. This improvement is further enhanced when historical kinematic features are also used. Interestingly, including historical kinematics improves alert timeliness more significantly than detection accuracy. This suggests that historical kinematics are important to be considered for earlier risk alerts.  

Across groups, the GSSMs that exclude and include the subject vehicle's instantaneous acceleration are contrasted. As is seen in Figure~\ref{fig: Result3}, including acceleration does not make improvements, except for $\mathrm{AUPRC}$ when historical kinematics are considered. This implies that the acceleration provides little additional benefit, probably because it is redundant with the historical kinematics, and is also affected by measurement noise and smoothing during trajectory processing.

\subsection{One GSSM for all interactions}\label{sec: exp4}
Driven by naturalistic data, GSSM is expected to learn interaction patterns across scenarios and effectively alert to potential collisions in different types of interactions. The crashes and near-crashes we processed from SHRP2 NDS are distributed in various types of events, as previously shown in Figure~\ref{fig: SHRP2_event_counts}. Accordingly, we compare the performance of GSSM and other 2D SSMs in these different safety-critical events. For better readability, we provide Table~\ref{tab: Result4} showing evaluation metrics, while Appendix Figure~\ref{fig: Result4} presents the performance curves. Table~\ref{tab: Result4} clearly shows that GSSM is significantly more effective in safety-focused detection accuracy across all interaction scenarios. This is particularly prominent for the conflicts beyond rear-end. Although the timeliness of GSSM is not always the best, it secures at least 1.88 seconds to prevent potential collisions in more than 75\% cases.
\begin{table}[htbp]
\centering
\caption{\textbf{Comparison of detection accuracy and alert timeliness in separate types of safety-critical events}. The GSSM under comparison is trained on the SafeBaseline dataset and uses contextual information of instantaneous motion kinematics, environmental conditions, and historical kinematics in the past 2.5 seconds. The 99\% confidence intervals (CI) of $m\mathrm{TTI}^*$ are obtained via two-sided sign test. For each type of event, the best value in each column is underlined and marked bold; the second-best value is marked bold.}
\label{tab: Result4}
\resizebox{\textwidth}{!}{%
\begin{tabular}{@{}lclccccccc@{}}
\toprule
\textbf{Event} & \textbf{\begin{tabular}[c]{@{}c@{}}Number\\of events\end{tabular}} & \textbf{Method} & $\mathrm{AUPRC}$ & $A_{80\%}^\mathrm{ROC}$ & $A_{90\%}^\mathrm{ROC}$ & $\mathrm{Precision}_{80\%}^\mathrm{PRC}$ & $\mathrm{Precision}_{90\%}^\mathrm{PRC}$ & $P^*_{\mathrm{TTI}\geq1.5}$ & $m\mathrm{TTI}^*$ $[Q1, Q3]$; $99\%CI$ \\
\midrule
\multirow{5}{*}{Rear-end} & \multirow{5}{*}{1787} & GSSM & 0.858 & \textbf{\underline{0.926}} & \textbf{\underline{0.887}} & \textbf{\underline{0.952}} & \textbf{\underline{0.930}} & \textbf{\underline{0.877}} & \textbf{\underline{2.53}} [1.92, 3.45]; 2.45--2.58 \\
 &  & ACT & \textbf{0.869} & \textbf{0.852} & \textbf{0.786} & 0.881 & 0.877 & 0.719 & 2.28 [1.30, 3.09]; 2.19--2.34 \\
 &  & TTC2D & \textbf{0.869} & 0.702 & 0.452 & \textbf{0.932} & \textbf{0.906} & \textbf{0.856} & \textbf{2.49} [1.85, 3.26]; 2.43--2.56 \\
 &  & TAdv & 0.726 & 0.734 & 0.636 & 0.798 & 0.749 & 0.486 & 1.44 [0.76, 2.04]; 1.39--1.50 \\
 &  & EI & \textbf{\underline{0.911}} & 0.745 & 0.562 & 0.930 & 0.812 & 0.730 & 2.10 [1.45, 2.88]; 2.02--2.16 \\
\cmidrule{1-10}
\multirow{5}{*}{Adjacent lane} & \multirow{5}{*}{611} & GSSM & \textbf{\underline{0.771}} & \textbf{\underline{0.693}} & \textbf{\underline{0.608}} & \textbf{\underline{0.702}} & \textbf{\underline{0.658}} & 0.826 & 3.38 [1.88, 5.15]; 3.19--3.61 \\
 &  & ACT & 0.685 & \textbf{0.532} & \textbf{0.334} & \textbf{0.683} & \textbf{0.608} & \textbf{\underline{0.896}} & \textbf{\underline{4.67}} [2.88, 6.14]; 4.40--4.94 \\
 &  & TTC2D & \textbf{0.698} & 0.317 & 0.000 & 0.632 & N/A & 0.848 & \textbf{3.68} [2.06, 5.38]; 3.36--3.85 \\
 &  & TAdv & 0.588 & 0.381 & 0.271 & 0.512 & 0.475 & 0.751 & 2.38 [1.45, 3.34]; 2.19--2.51 \\
 &  & EI & 0.636 & 0.220 & 0.042 & 0.451 & 0.451 & \textbf{0.851} & 3.48 [1.93, 5.50]; 3.33--3.74 \\
\cmidrule{1-10}
\multirow{5}{*}{Crossing/turning} & \multirow{5}{*}{93} & GSSM & \textbf{\underline{0.747}} & \textbf{\underline{0.599}} & \textbf{\underline{0.530}} & \textbf{0.755} & \textbf{\underline{0.729}} & \textbf{\underline{0.967}} & \textbf{4.85} [3.63, 6.38]; 4.21--5.46 \\
 &  & ACT & \textbf{0.735} & 0.350 & 0.017 & \textbf{\underline{0.792}} & \textbf{0.677} & 0.933 & \textbf{\underline{5.16}} [3.48, 6.46]; 4.62--5.40 \\
 &  & TTC2D & 0.678 & 0.000 & 0.000 & N/A & N/A & \textbf{0.950} & 4.77 [3.21, 6.33]; 4.32--5.31 \\
 &  & TAdv & 0.714 & \textbf{0.366} & \textbf{0.182} & 0.731 & 0.649 & 0.820 & 2.90 [2.04, 4.07]; 2.39--3.29 \\
 &  & EI & 0.694 & 0.000 & 0.000 & N/A & N/A & 0.934 & 4.13 [2.66, 5.71]; 3.30--4.70 \\
\cmidrule{1-10}
\multirow{5}{*}{Merging} & \multirow{5}{*}{29} & GSSM & \textbf{\underline{0.865}} & \textbf{\underline{0.795}} & \textbf{\underline{0.719}} & \textbf{\underline{0.828}} & \textbf{\underline{0.771}} & 0.889 & \textbf{5.11} [3.23, 7.30]; 3.58--5.60 \\
 &  & ACT & 0.705 & \textbf{0.520} & \textbf{0.225} & \textbf{0.758} & \textbf{0.659} & \textbf{\underline{1.000}} & \textbf{\underline{5.17}} [3.75, 6.80]; 4.52--6.49 \\
 &  & TTC2D & \textbf{0.786} & 0.000 & 0.000 & N/A & N/A & 0.857 & 3.83 [2.17, 4.85]; 2.66--4.23 \\
 &  & TAdv & 0.678 & 0.251 & 0.000 & 0.522 & N/A & 0.727 & 2.71 [1.32, 3.87]; 1.52--3.39 \\
 &  & EI & 0.777 & 0.286 & 0.000 & 0.553 & N/A & \textbf{0.897} & 3.21 [2.12, 4.85]; 2.52--4.43 \\
\cmidrule{1-10}
\multirow{5}{*}{\begin{tabular}[c]{@{}l@{}}With pedestrian/\\cyclist/animal\end{tabular}} & \multirow{5}{*}{31} & GSSM & \textbf{\underline{0.851}} & \textbf{\underline{0.379}} & \textbf{\underline{0.313}} & \textbf{\underline{0.800}} & \textbf{\underline{0.800}} & \textbf{\underline{1.000}} & \textbf{\underline{5.94}} [4.89, 7.16]; 4.89--7.16 \\
 &  & ACT & 0.706 & 0.066 & 0.000 & \textbf{0.758} & N/A & \textbf{0.966} & \textbf{5.38} [3.90, 6.54]; 4.20--6.09 \\
 &  & TTC2D & 0.697 & 0.000 & 0.000 & N/A & N/A & 0.897 & 4.51 [2.31, 5.57]; 2.82--5.14 \\
 &  & TAdv & 0.705 & \textbf{0.090} & \textbf{0.005} & 0.714 & \textbf{0.700} & 0.806 & 3.19 [1.76, 5.34]; 2.39--4.62 \\
 &  & EI & \textbf{0.733} & 0.000 & 0.000 & N/A & N/A & 0.867 & 3.78 [2.48, 5.55]; 2.55--5.36 \\
\bottomrule
\end{tabular}}
\end{table}

An interesting observation from Table~\ref{tab: Result4} is the superior performance of GSSM in rear-end and merging interactions. These two scenarios of interaction seem to be less challenging than other lateral interactions such as crossing and turning. Based on the performance, the most challenging scenarios are interactions with pedestrians, cyclists, and animals. This could be because of the less predictable behaviour of these active and flexible road users, but also could be because the interaction patterns with them are very limitedly covered in the training data used in this paper.

\subsection{Attribution of collision risk}\label{sec: exp5}
\begin{figure}[htbp]
    \centering
    \includegraphics[width=\linewidth]{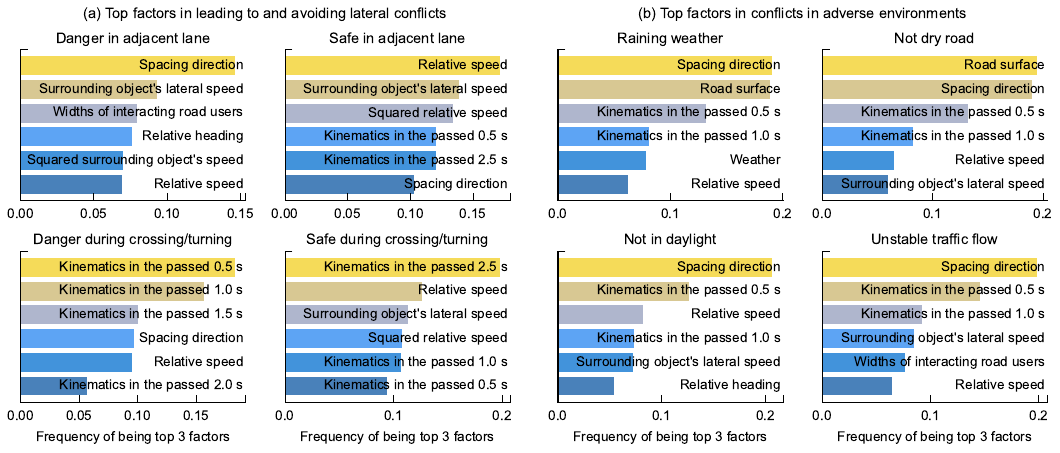}
    \captionof{figure}{Top-ranked factors in risk quantification by GSSM. The GSSM used for feature attribution is S-CET, which is trained on SafeBaseline and uses current, environment, and historical kinematic features. The threshold used to distinguish danger and safety is 2.52, which is optimised to achieve the highest $\mathrm{Precision}$ for $\mathrm{Recall\geq0.85}$. This threshold means an interaction is considered unsafe if its spacing remains minimal for at least $10^{2.52}\approx331$ times of observations in the same interaction context. \textbf{(a)} Top-ranked factors in leading to and avoiding lateral crashes and near-crashes. \textbf{(b)} Top-ranked factors in crashes and near-crashes happened in adverse environments.}
    \label{fig: Result5}
\end{figure}

Figure~\ref{fig: Result5} presents the top-ranked factors in different situations when GSSM evaluates an interaction to be safe or not, which are respectively considered in safe periods and danger periods as defined in Figure~\ref{fig: Safe_danger_period}. At each time moment in a considered period, we calculate the attributions of all features, compare the positive attributions if the considered period is safe or negative attributions otherwise, and record the 3 factors with the highest attributions. Then the ranking of the most contributing factors is obtained after comparing the features at all time moments. Note that because our feature attribution is performed based on the encoded representation rather than raw numbers of variables, the summarised attributions indicate relative importance without implying whether an increasing or decreasing value is associated with the quantified risk.

Focusing on what leads to potential collisions in lateral interactions, the left half of Figure~\ref{fig: Result5}(a) shows that spacing direction, i.e., the angular coordinate of multi-direction spacing as defined in Section~\ref{sec: 2dspacing}, contributes the most to dangers in the adjacent lane, while the historical kinematics in the past 1.5 seconds contribute the most to dangers during crossing and turning. Spacing direction is related to relative heading and the surrounding object's lateral speed, which are also ranked high. With not much surprise, relative speed also contributes considerably to potential collision. In contrast, the right half of Figure~\ref{fig: Result5}(a) shows factors contributing to safe interactions. Relative speed and the surrounding object's lateral speed are the most contributing ones. In addition to them, relatively longer historical kinematics are important. These results suggest that relative direction and speed are the main factors for accurate risk quantification. Notably, a short movement history of approximately 1.5 seconds is more influential in detecting danger, whereas a longer history of around 2.5 seconds provides greater context for perceiving safety in lateral interactions.

Adverse environments such as precipitation, wet road surfaces, driving at night, or unstable traffic flow are relatively infrequent conditions, but may significantly increase the risk of collisions. In Figure~\ref{fig: Result5}(b), we present the factors contributing to dangers in different adverse environments. The top factors harming safety overlap with rain and when the road surface is wet. In both conditions, spacing direction and road surface are ranked as the most important. This implies that the collision risk in such situations is highly influenced by the directional control of the involved road users. Spacing direction remains highly ranked when considering interactions in the dark as well as in unstable traffic flow. The other factors in these two adverse environments resemble those in general interactions in Figure~\ref{fig: Result5}(a). Interestingly, the widths of interacting road users, i.e., half of the sum of the vehicle widths, play a role in collision risk when the traffic flow is unstable. This could be due to relatively restrained road space for safe interactions.

\section{Conclusion and discussion}\label{sec: conclusion}
We present the generalised surrogate safety measure (GSSM) to proactively quantify the risk of potential traffic collisions before they happen. GSSM intends to address the challenges of scalability, context-awareness, and generalisability, faced by existing approaches to proactive collision risk quantification. Below, we discuss the main findings.
\begin{itemize}[noitemsep,topsep=-\parskip]
    \item Instead of relying on historical records of crashes or near-crashes, GSSM stably learns from the patterns of normal interactions and extrapolates them to safety-critical situations. This allows for \textbf{\emph{scalable}} improvement with increasing amounts of data, as are being collected by automated vehicles, and our results show enhanced accuracy in risk quantification by additional lateral interaction patterns. 
    \item The fundamental assumption made by GSSM is that collision risk emerges if interactions become extreme, which is measured by the spacing between road users involved in a specific interaction context. GSSM utilises neural networks to approximate such context-conditioned distributions of multi-directional spacing and is data-driven. Its risk quantification is thus \textbf{\emph{context-aware}} by incorporating any potentially helpful information as inputs, e.g., instantaneous motion states, weather, road surface condition, lighting, and historical kinematics, as we have illustrated. 
    \item Our results also demonstrate that GSSM is highly \textbf{\emph{generalisable}}. It achieves superior accuracy and timeliness in alerting potential collisions across various interaction scenarios such as rear-end, merging, crossing, turning, etc. Impressively but reasonably, GSSM also generalises well from training data to unseen test data, even when the training data are collected in different countries and by different equipment. This implies that the interaction patterns in urgent conditions, such as crashes and near-crashes, are shared in human behaviour.
    \item The attribution of collision risk shows that GSSM correctly utilises the information of weather and road surface conditions when it is rainy or when the road is not dry. Importantly, \textbf{\emph{spacing direction}} has a dominant influence in lateral interactions, where short motion histories of around 1.5 seconds help recognise risk and longer motion histories of 2.5 seconds are more useful to confirm safety. These findings emphasise the necessity for context-aware collision risk quantification to consider diverse factors.
\end{itemize}
\vspace{\parskip}

GSSM is limited by our modelling choices and requires further development and targeted validation in the future. First of all, this study focuses on the probability of potential collisions while omitting to consider collision severity or time of occurrence, which are important aspects of collision risk. Driven by real-world data, GSSM's robustness to sensor noise, missing values, and small perturbations in the input is not guaranteed, yet our training does not provide formal guarantees or certified bounds. The datasets used in this research are carefully processed. To verify GSSM’s stability and transferability, especially when extrapolating to extreme behaviours, a larger variety of interaction data with multi-source noise and from more different driving cultures will be essential. In addition, richer modalities such as vision and language can be incorporated to characterise more informative contexts. This motivates representation learning to encode complex context and enable large-scale training in industrial applications. To truly improve traffic safety, it is also important to investigate not only correlations but also causal mechanisms between contextual features and the estimated risk. We regard GSSM as a starting point toward foundation models for proactive risk quantification of potential collisions. Before achieving that goal, more real-world training and verification are required to confirm its stable effectiveness, quantifiable scalability, explainable context-awareness, and controllable generalisability. 

The implications of such a methodology are broad for both ex ante prevention and ex post analysis of traffic collision risk. The most direct benefit is for autonomous driving systems (ADS), given that GSSM can compute for a thousand interactions per query within 25 milliseconds. Online, GSSM can provide a time-varying risk signal that serves as a safety cost or constraint in motion planning, or as a safety ``shield’’ that warns of potentially unsafe behaviours before they are executed. Offline, it can be used to train and validate ADS in safety-critical situations, e.g., by filtering valuable training materials, generating or ranking validation scenarios at varying levels of collision risk, and supporting closed-loop testing in which high-risk interactions are selectively evaluated. Beyond ADS, GSSM enables civil engineering and traffic management use cases: it allows for proactive evaluation and thus data-driven improvement in the safety of road designs, operations, and policies. For example, GSSM can help traffic engineers and policy makers detect emerging hazards in road networks and mitigate potential accidents before they occur. Altogether, these applications contribute to a significant step toward the long-term vision of zero traffic fatalities, via a tool that quantifies and manages risk proactively, in any context, at any time.

\section*{Data availability}\label{sec: data availability}
The raw data are sourced from two subsets of the SHRP2 Naturalistic Driving Study: ``A Study on the Factors That Affect the Occurrence of Crashes and Near-Crashes''~\cite{Sears2019Honda} and ``Research of Driver Assistant System''~\cite{Layman2019DAS}. They can be accessed following the instructions at {\small\url{https://github.com/Yiru-Jiao/GSSM}}. All resulting data in this study are openly accessible via 4TU.ResearchData at {\small\url{https://doi.org/10.4121/9caa1e6c-9abd-4e36-ae28-c9ea4542d940}}~\cite{data_repo}.

\section*{Code availability}\label{sec: code availability}
The code for this research is open-sourced at {\small\url{https://github.com/Yiru-Jiao/GSSM}}. A permanent record of the code repository by this publication is managed by Zenodo at {\small\url{https://doi.org/10.5281/zenodo.17099863}}~\cite{github_repo}.

\section*{Acknowledgments}
This work is supported by the TU Delft AI Labs programme. The authors acknowledge the use of computational resources of the DelftBlue supercomputer, provided by Delft High Performance Computing Centre ({\small\url{https://www.tudelft.nl/dhpc}}).

We extend our sincere gratitude to the researchers and organisations who collected, created, cleaned, and curated the high-quality datasets for research use. We thank Dr. Guopeng Li for his knowledge in neural network training.

The raw data were accessed under SHRP2 Data Use License SHRP2-DUL-A-5-24-746, issued by the Virginia Tech Transportation Institute (VTTI). The findings and conclusions presented here are those of the authors and do not necessarily represent the views of VTTI, the Transportation Research Board, the National Academies, or the Federal Highway Administration.

\section*{Declaration of generative AI and AI-assisted technologies in the writing process}
During the preparation of this work, the authors used ChatGPT and DeepSeek in order to obtain suggestions for readability improvement. No sentence was entirely generated by the generative tools. After using the tools, the authors have reviewed and edited the content as needed and take full responsibility for the content of the publication.

\printbibliography

\newpage
\begin{appendices}

\section*{Appendix}
\renewcommand{\thesubsection}{A.\arabic{subsection}}
\renewcommand\thefigure{A.\arabic{figure}}
\renewcommand\thetable{A.\arabic{table}}
\renewcommand{\theequation}{A.\arabic{equation}}
\setcounter{figure}{0}
\setcounter{equation}{0}
\setcounter{table}{0}
\subsection{SHRP2 trajectory reconstruction error}
\begin{figure}[htbp]
    \centering
    \includegraphics[width=\linewidth]{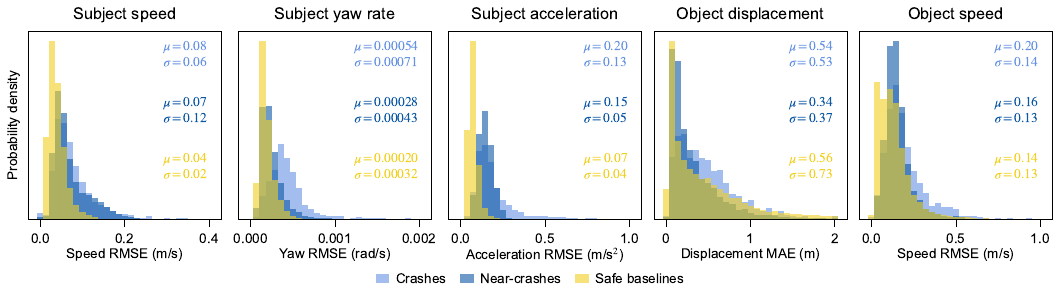}
    \caption{Reconstruction error distributions of crashes, near crashes, and safe baselines in the SHRP2 NDS. The mean values ($\mu$) and standard deviations ($\sigma$) are marked in each plot.}
    \label{fig: SHRP2_error_distributions}
\end{figure}

Figure~\ref{fig: SHRP2_error_distributions} shows the error distributions of bird's eye view reconstruction of the trajectories derived in the Second Strategic Highway Research Program’s (SHRP2) Naturalistic Driving Study (NDS). These include the root mean squared error (RMSE) in subject speed, subject yaw rate, subject acceleration, and object speed, as well as the mean absolute error (MAE) of object displacement. The errors are evaluated for each event in the categories of crashes, near-crashes, and safe baselines.

\subsection{Contextual information details}
As explained in Section~\ref{sec: context rl}, this study considers 3 categories of contextual information: current features ($X_C$) including the instantaneous states of interacting road users, categorical environment features ($X_E$) of external conditions during the interaction, and historical kinematic features ($X_T$) composed of time-series speeds and yaw rates within the past 2.5 seconds. Table~\ref{tab: context features} provides a detailed list of these contextual features used in this study, as well as their encoding unit.
\begin{table}[htb]
\centering
\caption{Contextual information considered in this study.}
\label{tab: context features}
{\small
\begin{tabular}{@{}cll@{}}
\toprule
\textbf{Category} & \textbf{Attribution unit} & \textbf{Feature(s)} \\ \midrule
$X_C$ & Per each feature& $ \{l_i, l_j, \frac{w_i+w_j}{2}, |\boldsymbol{v}_i|, x_{\boldsymbol{v}_j}, y_{\boldsymbol{v}_j}, |\boldsymbol{v}_i|^2, |\boldsymbol{v}_j|^2, |\boldsymbol{v}_{ij}|^2, |\boldsymbol{v}_{ij}|\mathrm{sgn}(\Delta v),A_{\boldsymbol{h}_j},\rho_{ij}\}_{t=0}$, optional $a_{i,t=0}$ \\ \cmidrule{1-3}
\multirow{30}{*}{$X_E$} & \multirow{6}{*}{Lighting condition} & Darkeness (lighted), \\
 & & darkness (not lighted), \\
 & & dawn, \\
 & & daylight, \\
 & & dusk, \\
 & & unknown. \\ \cmidrule{2-3}
 & \multirow{8}{*}{Weather condition} & No adverse conditions, \\
 & & fog, \\
 & & mist/light rain, \\
 & & rain and fog, \\
 & & raining, sleeting, \\
 & & snow/sleet and fog, \\
 & & snowing, \\
 & & unknown. \\ \cmidrule{2-3}
 & \multirow{8}{*}{Road surface} & Dry, \\
 & & gravel over asphalt, \\
 & & gravel/dirt road, \\
 & & icy, \\
 & & muddy, \\
 & & snowy, \\
 & & wet, \\
 & & unknown. \\ \cmidrule{2-3}
 & \multirow{8}{*}{Traffic density} & LOS A1: free flow, no lead traffic, \\
 & & LOS A2: free flow, leading traffic present, \\
 & & LOS B: flow with some restrictions, \\
 & & LOS C: stable flow, manoeuvrability and speed are more restricted, \\
 & & LOS D: unstable flow - temporary restrictions substantially slow driver, \\
 & & LOS E: flow is unstable, vehicles are unable to pass, temporary stoppages, etc., \\
 & & LOS F: forced traffic flow with low speeds and traffic volumes that are below capacity, \\
 & & unknown. \\ \cmidrule{1-3}
\multirow{5}{*}{$X_T$} & In the passed 0.5 s & $\{\omega_i, |\boldsymbol{v}_i|, x_{\boldsymbol{v}_j}, y_{\boldsymbol{v}_j}\}_{t\in\{-0.5,-0.4,\dots,-0.1\}}$ \\
 & In the passed 1 s & $\{\omega_i, |\boldsymbol{v}_i|, x_{\boldsymbol{v}_j}, y_{\boldsymbol{v}_j}\}_{t\in\{-1.0,-0.9,\dots,-0.1\}}$ \\
 & In the passed 1.5 s & $\{\omega_i, |\boldsymbol{v}_i|, x_{\boldsymbol{v}_j}, y_{\boldsymbol{v}_j}\}_{t\in\{-1.5,-1.4,\dots,-0.1\}}$ \\
 & In the passed 2 s & $\{\omega_i, |\boldsymbol{v}_i|, x_{\boldsymbol{v}_j}, y_{\boldsymbol{v}_j}\}_{t\in\{-2.0,-1.9,\dots,-0.1\}}$ \\
 & In the passed 2.5 s & $\{\omega_i, |\boldsymbol{v}_i|, x_{\boldsymbol{v}_j}, y_{\boldsymbol{v}_j}\}_{t\in\{-2.5,-2.4,\dots,-0.1\}}$ \\ \bottomrule
 & & \\
\multicolumn{3}{@{}p{\linewidth}@{}}{\textbf{Note}: for two interacting road users $i$ and $j$, their yaw rates, velocities, lengths, widths, and headings are denoted by $\omega\in[-\pi,\pi]$, $\boldsymbol{v}$, $l$, $w$, and $\boldsymbol{h}$; the motion states are in a local coordinate system with the origin at the position of $i$ and y-axis oriented along the velocity direction of $i$; $\mathrm{sgn}(\Delta v)=\mathrm{sgn}(|\boldsymbol{v}_i|-|\boldsymbol{v}_j|)$; $A_{\boldsymbol{h}}\in[-\pi,\pi]$ denotes the angle between the heading direction and the y-axis; $\rho_{ij}\in[-\pi,\pi]$ is the angle in the polar coordinates of multi-directional spacing.}\\
\end{tabular}}
\end{table}

\subsection{Neural network architecture}\label{sec: nn architecture}
Out of the aim of validating the generalised surrogate safety measure (GSSM) rather than looking for the most powerful architecture, our neural networks are relatively simple for computational convenience. As mentioned in Section~\ref{sec: context rl}, we have 3 encoders and 1 decoder. The encoders for current features ($X_C$) and environment features ($X_E$) are multi-layer perceptrons (MLPs,~\cite{mlp}), while the encoder for historical kinematic features ($X_T$) uses a single-layer long short-term memory (LSTM,~\cite{lstm}) recurrent neural network. The encoded representation groups are then concatenated and passed to the decoder, which uses the attention mechanism~\cite{attention} to capture inter-feature relations and convolutional neural networks (CNN,~\cite{cnn}) for intra-feature relations. Throughout the model, we use the activation function of Gaussian error linear units (GELU,~\cite{gelu}). More descriptive details are as follows, while implementation details are referred to in our open-sourced repository.

To ensure that feature attributions are meaningful, we design the encoders to let each attribution unit (a representation vector) carry independent information from the others. Aligned with Table~\ref{tab: context features}, each scalar feature in $X_C$ is mapped to its own representation vector. The categorical features in $X_E$ are first one-hot encoded and grouped into four chunks (weather, lighting, road surface, and traffic density), and each chunk is then mapped to a separate representation vector. For $X_C$ and $X_E$, we use MLP encoders with 5 and 4 linear layers, respectively. The temporal features $X_T$ are encoded into five representation vectors corresponding to the past 0.5, 1, 1.5, 2, and 2.5 seconds. Concretely, we reverse the time steps in the 2.5-second history and feed the reversed sequence into a single-layer unidirectional LSTM. From the resulting output sequence, we take five representations at 0.5-second intervals, so that each one encodes the kinematic features within 0.5, 1, 1.5, 2, and 2.5 seconds closest to the current moment. 

To enrich and regularise the latent representation, we append a set of orthogonal Gaussian random features~\cite{random_features} which are deterministic and unique for a sequence of representation vectors. Seeing the encoded and enriched vectors as tokens of different contextual information, we design the decoder to capture both global and local relations of the tokens. We first use batch normalisation to align the ranges of all feature dimensions, and then stack 6 self-attention blocks, in each of which the feed-forward network has 2 linear layers. Next, we apply a CNN with 2 layers of convolution and a kernel size of 3. Lastly, we use two separate 3-layer MLPs to output the targeted parameters $\mu$ and $\log(\sigma^2)$. 

\subsection{Hyperparameter settings}\label{sec: hyperparameters}
The performance of GSSM depends on a number of implementation choices such as learning-rate schedules, batch sizes, and regularisation strengths. To ensure transparency and reproducibility, we report a full set of hyperparameters used in the experiments in Table~\ref{tab: hyperparameters}.
\begin{table}[htbp]
\centering
\caption{Hyperparameters set in this paper's experiments.}
\label{tab: hyperparameters}
\begin{tabular}{@{}lll@{}}
\toprule
\textbf{Hyperparameter} & \textbf{Definition} & \textbf{Value} \\ \midrule
Random seed & Number that controls generation of pseudo-random values & 131 \\
Small value threshold & Threshold to consider a value smaller than it to be near zero & 1e-6 \\
Representation dimension & Number of variables in a piece of encoded representation & 64 \\
Perturbation noise & Standard deviation of the added Gaussian noise & 1\% of variable range \\
Max. epochs & Maximum epochs of training & 150 \\
Batch size & Number of samples used per update during model training & 512 \\
Initial learning rate & Learning rate at the beginning of model training & 1e-4 \\
Dropout & Probability for each value in a tensor to be set to 0 & 0.2 \\
\bottomrule
\end{tabular}
\end{table}

\subsection{Computation efficiency}\label{sec: computation efficiency}
As reported in Table~\ref{tab: gssms2learn}, a GSSM has 773,286 parameters if the input context has $X_C$ only; when both $X_C$ and $X_E$ are considered, there are 857,486 parameters; if $X_T$ is also included, the number of parameters becomes 885,664. The dominant part of the parameters is of the decoder, correspondingly accounting for 95.92\%, 87.46\%, and 85.83\% of all parameters in a model. Such a model with fewer than 0.9 million parameters is fairly small for modern deep learning. 

To quantify the computational cost, we further report the training time and inference time for the different GSSM variants alongside other existing methods for comparison. Note that we implement TTC2D, ACT and TAdv in a vectorised form for efficient large-scale evaluation. In contrast, EI is used as provided by its authors, as its formulation does not readily admit the same level of vectorisation. Altogether, these results show that the proposed GSSM achieves competitive computational efficiency while providing substantially richer, context-aware risk estimates.

\subsection{Performance curves in different interactions}
\begin{figure}[htb]
    \centering
    \includegraphics[width=\linewidth]{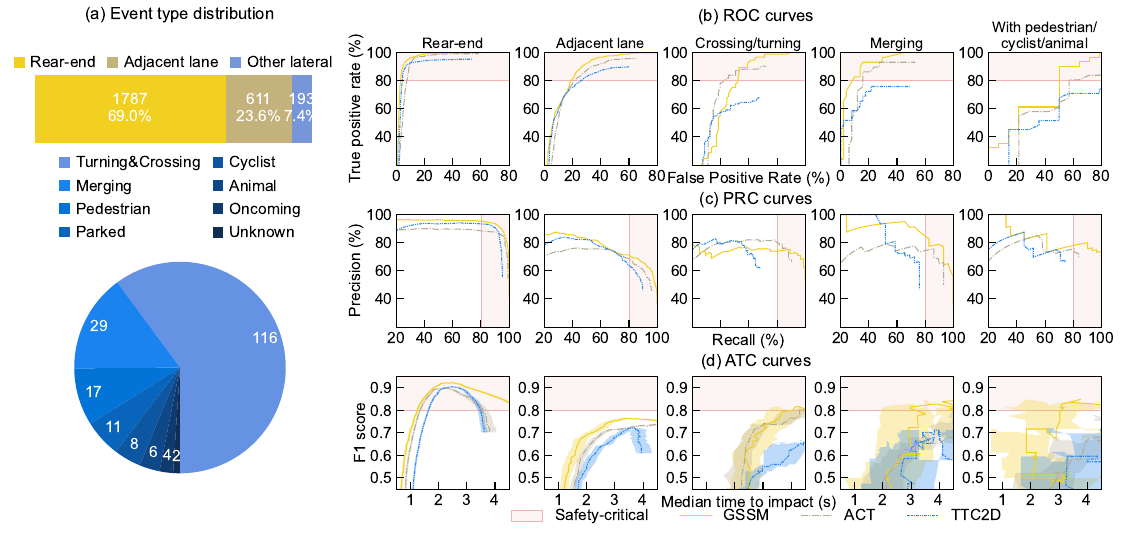}
    \captionof{figure}{Performance comparison of GSSM and other 2D SSMs in alerting different types of safety-critical events. The GSSM under comparison is S-CET, which is trained on SafeBaseline and uses contextual information of current features, environment features, and historical kinematic features. \textbf{(a)}~Types of the crashes and near-crashes with determined ground truth. \textbf{(b)}~Receiver operating characteristic curves for different types of events. \textbf{(c)}~Precision-recall curves for different types of events. \textbf{(d)}~Accuracy-timeliness curves for different types of events.}
    \label{fig: Result4}
\end{figure}

Corresponding to Table~\ref{tab: Result4}, Figure~\ref{fig: Result4} shows the performance curves to compare GSSM and other two-dimensional surrogate safety measures (2D SSMs) in different types of interactions. The curves are less smooth for interactions of merging and with pedestrian/cyclist/animal because of the small number of test events.

\end{appendices}
\end{document}